\documentclass[11pt]{article}

\usepackage[preprint]{acl}

\usepackage{times}
\usepackage{latexsym}
\usepackage[inline]{enumitem} 

\usepackage[T1]{fontenc}

\usepackage[utf8]{inputenc}

\usepackage{microtype}

\usepackage{inconsolata}

\usepackage{graphicx}
\usepackage{booktabs}
\usepackage{amsmath}
\usepackage{amsthm}
\usepackage{multirow}
\usepackage{comment}

\newcommand{\bocr}[1]{\textcolor{black}{#1}}
\newcommand{\boarxiv}[1]{\textcolor{black}{#1}}
\newcommand{\bo}[1]{\textcolor{black}{#1}}
\newcommand{\tc}[1]{\textcolor{black}{#1}}
\newcommand{\dc}[1]{\textcolor{black}{#1}}
\newcommand{\tcarr}[1]{\textcolor{black}{#1}}

\usepackage{tcolorbox}
\usepackage{soul}
\newcommand{\ctext}[3][RGB]{%
  \begingroup
  \definecolor{hlcolor}{#1}{#2}\sethlcolor{hlcolor}%
  \hl{#3}%
  \endgroup
}

\title{When Hate Meets Facts: \\ LLMs-in-the-Loop for Check-worthiness Detection in Hate Speech}

\author{
  \textbf{Nicolás Benjamín Ocampo\textsuperscript{1}}, 
  \textbf{Tommaso Caselli\textsuperscript{2}}, 
  \textbf{Davide Ceolin\textsuperscript{1}}
\\
  \textsuperscript{1}HCDA, Centrum Wiskunde \& Informatica, Amsterdam, The Netherlands \\
  \textsuperscript{2}CLCG, University of Groningen, Groningen, The Netherlands \\
   \small{\texttt{n.b.ocampo@cwi.nl, t.caselli@rug.nl, d.ceolin@cwi.nl}}
}

\begin{document}
\maketitle
\begin{abstract}

Hateful content online is often \tcarr{expressed} using fact-like, not necessarily correct information, especially in coordinated online harassment campaigns and extremist propaganda. Failing to jointly address hate speech (HS) and misinformation can deepen prejudice, reinforce harmful stereotypes, and expose bystanders to psychological distress, while polluting public debate. Moreover, \tcarr{these messages require more effort from content moderators because they must assess both harmfulness and \tcarr{veracity}, i.e., fact-check them.} 
To address this challenge, \bo{we release WSF-ARG+, the first dataset \tcarr{which combines hate speech with check-worthiness information}.} We also \tcarr{introduce a novel LLM-in-the-loop framework to facilitate the annotation of check-worthy claims.} 
\bo{We run our framework\dc{, testing it with} 12 open-weight LLMs of different sizes and architectures. We validate it through extensive human evaluation, \dc{and show} that our LLM-in-the-loop framework reduces human effort without compromising the annotation quality of the data.}
\bo{Finally, we show that HS messages with check-worthy claims show significantly higher harassment and hate, and that incorporating check-worthiness labels improves LLM-based HS detection \tcarr{up to 0.213 macro-F1} and  to 0.154 macro-F1 on average for large models.} 
\end{abstract}

{\color{red} \noindent\textbf{Content Warning}: This paper contains examples of language that may be offensive to some readers.}

\section{Introduction}

The quality of content \tcarr{in the online sphere} has been \tcarr{constantly} decreasing~\citep{doctorow2025enshittification,shaib2025measuringaisloptext}. \tcarr{The large amount of user-generated content and the increasing laxity of content moderation policies on social media platforms} 
have facilitated the spread of hate speech and misinformation, posing a global threat that undermines efforts \tc{to maintain healthy online environments} and stable democracies~\citep{vanDerLinden2023_APA_misinformation}. 

Previous work has investigated hate speech and misinformation as separate phenomena. On the other hand, following~\citet{WardleDerakhshan2017}, hate speech (HS) and misinformation are strictly related~\cite{10.1093/pnasnexus/pgae111}, and they represent two instances of a more general phenomenon labeled \textit{information disorder}.
Evidence of the strict connection between these instances 
can be found in recent tragic events such as the Rwanda and Rohingya genocides: in both cases, as attested by UN reports,\footnote{Rwanda: \url{https://bit.ly/3NsiCdp}; Rohingya: \url{https://bit.ly/4pt7jyQ}} misinformation was used to promote hate against specific ethnic groups who have been persecuted.


From a fact-checking perspective, misinformation—defined as demonstrably false or misleading information, regardless of source or intent—is linguistically realized through \textit{claims}. Claims are statements that express assertions on a specific topic and represent the author's commitment to presenting a proposition as fact~\citep{murzaku2022re}. \bocr{In hate speech—defined as a deliberate attack directed toward a specific group of people and motivated by aspects of that group’s identity \cite{de-gibert-etal-2018-hate}—
the attack (or components of it) can be realized via claims that contain demonstrably false information (e.g., \textit{\textbf{``The kind of perversion that leads to homosexuality also leads to pedophilia. Gay men may turn out to be pedophiles.}''}) or truthful information deployed misleadingly (e.g., \textit{\textbf{``Africa is an area rich in natural resources but they only use these resources to make mud huts}.''}), and thus reinforcing the hateful stances.}

Identifying whether a claim warrants fact-checking, i.e., establishing its check-worthiness, is therefore an essential step that enables the assignment of truthfulness values to claims, facilitating targeted interventions that disrupt the misinformation-hate nexus. By prioritizing fact-checking resources on claims that meaningfully contribute to hateful narratives, it becomes possible to respond to hate with appropriate instances of counterspeech. This will debunk the
false premises that legitimize prejudice while educating audiences about manipulative rhetorical strategies. To this end, we propose the following contributions: 
\bo{
\begin{itemize}
\item We \tcarr{introduce} a new dataset, WSF-ARG+, which enrich WSF-ARG~\cite{bonaldi-etal-2024-safer} with non-hateful messages and check-worthiness annotations \tcarr{to support the study of interactions between HS and misinformation}.\footnote{The annotated data, and the accompanying annotation
guidelines and code can be found at \url{https://github.com/benjaminocampo/wsf_arg_plus}.}
\item We \tcarr{propose and validate} a novel LLM-in-the-loop framework \tcarr{to facilitate the annotation of check-worthy claims; we also present} strategies for mitigating errors through model selection and human oversight \tcarr{to address intrinsic limitations of LLMs as data annotators~\cite{baumann2025largelanguagemodelhacking}}. The framework enables check-worthiness annotation to reach full human-level quality. 
\item We \tcarr{investigate} the intersection between hate speech and check-worthiness in WSF-ARG+, \tcarr{showing that} 
messages containing check-worthy claims are 
more harmful and hateful \tcarr{and that} 
incorporating check-worthiness labels 
improves hate speech detection \tcarr{up to 0.213 macro-F1 and to 0.154 macro-F1 on average for large models. 
}
\end{itemize}
}


\section{Related Work}
\label{sec:related}

This section reviews how LLMs have been 
for annotating unstructured text (Section \ref{sec:llm-for-ir}). We then examine the overlap between hate speech and misinformation, noting that many datasets remain unrepresentative and often include hate without factual claims (Section \ref{sec:hs-misinf-datasets}). Finally, we situate check-worthiness detection within the broader misinformation-detection landscape (Section \ref{sec:checkworthiness_sota}).


\subsection{LLM as Data Annotators}
\label{sec:llm-for-ir}

LLMs have increasingly been adopted in research as a means to bootstrap relevant information, process data, or generate relevant labels on unstructured data. Prior research has shown that LLM-based annotations can substantially reduce labeling costs~\citep{wang-etal-2021-want-reduce} and, in some cases, even surpass the quality of crowd-sourced gold references, underscoring their growing relevance for computational social science~\citep{ziems-etal-2024-large,gpt-4-as-annotator}. 
\citet{zhu2025judgelmfinetunedlargelanguage} introduce JudgeLM, a fine-tuned model designed for open-ended evaluation tasks such as counterspeech assessment~\citep{zubiaga-etal-2024-llm,bonaldi-etal-2025-first}.~\citet{lin-etal-2023-argue} propose ArgJudge and ChatGPTEval, two LLM-based approaches to evaluate the relevance and informativeness of generated counter-arguments. 

However, concerns remain about the reliability and generalizability of LLM-based annotations. Model performance depends on task subjectivity~\citep{li-etal-2023-synthetic} and some models exhibit primacy bias, favoring labels presented earlier in prompts~\citep{wang-etal-2023-primacy}.~\citet{baumann2025largelanguagemodelhacking} demonstrate that implementation choices (e.g., model selection, prompting strategy) substantially affect annotation outcomes, introducing systematic and random errors. These findings underscore the need for transparency and methodological rigor when employing LLMs for annotation.

\subsection{Hate Speech and Misinformation Datasets}
\label{sec:hs-misinf-datasets}

Many widely used datasets in hate speech~\cite[among others]{Davidson2017,basile-etal-2019-semeval,zampieri-etal-2020-semeval}  
primarily consist of short, explicit \boarxiv{messages} not expressing claims and relying on slurs, insults, or other \tcarr{transparent} hateful terms, making them unsuitable to evaluate the check-worthiness of their claims.
Other datasets~\citep{sap-etal-2020-social,elsherief-etal-2021-latent,vidgen-etal-2021-learning,hartvigsen-etal-2022-toxigen,ocampo-etal-2023-depth} 
are better candidates \tcarr{but they still contain a significant proportion of short, explicit hateful statements, and many of their implicit hate messages are either brief or rely on coded language to obscure stance without providing supporting reasons~\citep{bonaldi-etal-2024-safer}} 

Datasets oriented toward counterspeech generation, such as CONAN~\citep{chung-etal-2019-conan} and MT-CONAN~\citep{fanton-etal-2021-human}, are designed to provide non-aggressive, \tc{fact-based} responses to hateful messages. However, even these datasets are \tc{partially generated in a synthetic way.} For instance, MT-CONAN is derived from CONAN through machine translation from English. Additionally, \bo{more than half of the hate messages in MT-CONAN are not check-worthy (see Appendix \ref{sec:cw-det-other-datasets}).} 
The importance of ``ecological'' data has motivated the development of WSF-ARG~\citep{bonaldi-etal-2024-safer}, a dataset with hateful arguments 
to generate argument-oriented counter-speech.

From the misinformation perspective, well-known datasets such as \bo{LIAR++ and FullFact}~\citep{russo-etal-2023-benchmarking}, and \tc{their} extended versions like Vermouth~\citep{russo-etal-2023-countering} and LIAR-ARG~\citep{Wang2025}, contain \bo{claims} originally sourced from fact-checking organizations. These \bo{claims} are check-worthy by design but lack hateful annotations. 

\subsection{Check-worthiness Detection}
\label{sec:checkworthiness_sota}

Most of the previous work in misinformation detection has focused on two tasks: claim detection and factuality assessment~\citep{guo-etal-2022-survey,hardalov-etal-2022-survey,zerong-etal-2025-systematic} 
While strictly connected to claim detection, check-worthiness detection has received less attention. \tcarr{One of the first datasets supporting both claim detection and check-worthiness detection is ClaimBuster~\citep{claimbuster}, which annotates presidential debate claims across three classes: non-factual, factual but not check-worthy, and factual and check-worthy. A major subsequent effort is the CheckThat! Lab\footnote{\url{https://checkthat.gitlab.io/editions/}} series at CLEF (2021--2024), which released multiple datasets varying in text type, modality, and language. Submitted systems fall into two categories: supervised fine-tuned models and, since 2023, prompt-based LLM methods, with the former consistently outperforming the latter.}

\tcarr{More recently,~\citet{majer-snajder-2024-claim} evaluated zero- and few-shot LLM prompting for check-worthiness detection across multiple datasets, systematically varying prompt verbosity and contextual information. They find that optimal verbosity is dataset-dependent, metadata yields greater gains than co-text, and well-calibrated confidence scores enable reliable ranking. Overall, the approach matches fine-tuned baselines on most datasets, demonstrating that annotation guidelines can effectively guide LLM-based detection. Building on these findings, we employ LLMs to bootstrap check-worthiness annotation for our data.}

\section{Methodology}
\label{sec:methodology}

\bo{In this section, we first define the check-worthiness labeling task (Section \ref{sec:cw-detection}). Second, we present our LLM-in-the-loop framework for annotating check-worthy claims (Section \ref{sec:llm-in-the-loop-pipeline}).} 

\subsection{Check-worthiness Task Description}
\label{sec:cw-detection}

\tc{Check-worthiness entails two dimensions: (i.) it requires messages to contain claims, i.e., verifiable assertions about the world; 
(ii.) claims must have a potential to have an impact or be of interest to the general public~\citep{alam-etal-2021-fighting-covid}.}
Check-worthy claims are likely to be noticed, repeated, or influential, such that a typical member of the public might reasonably ask whether they are true. Because they involve factual claims of public interest, these claims are especially relevant for journalists and fact-checkers seeking to inform the public \citep{claimbuster,jaradat-etal-2018-claimrank,wright-augenstein-2020-claim,ijcai2021p619}. We frame check-worthiness labeling as a multi-class annotation task based on the ClaimBuster framework: 

\paragraph{Checkworthy Factual Statement (CFS)} It contains a factual claim that is deemed important and should be fact-checked due to its impact on the public. This includes verifiable claims, viral or widely shared false information, hoaxes without evidence, and provably false claims that are timely, trending and consequential in their context (e.g., \textit{``92\% of abortion clinics are in black communities.''}).

\paragraph{Unimportant Factual Statement (UFS)} It contains a factual claim that is not important enough for fact-checking. These are claims of low 
interest to the general public (e.g., \textit{``\underline{When I was last in South Africa}, we went to Sun City and saw little black monkey children swimming.''}). 

\paragraph{Non-Factual Statement (NFS)} A subjective or opinion-based text that includes beliefs, personal declarations, wishes, or rhetorical questions. These do not contain factual claims (e.g., \textit{``We \underline{should go} to the southwest and go on the offensive and drive every race down into Mexico.''}). 

\subsection{LLM-in-the-loop: Bootstrapping Data Annotation Avoiding Over-reliance}
\label{sec:llm-in-the-loop-pipeline}
\tcarr{Annotation tasks require multiple annotators. Their number can vary according to various factors such as expertise, complexity of the task, and the annotation paradigm (prescriptive vs. descriptive)~\cite{rottger-etal-2022-two}. In the design of our LLM-in-the-loop framework, we have assumed the following settings: \dc{we recruit only expert annotators}, the guidelines are ``rigid'' (i.e., prescriptive), and there must be a final gold label. Under this perspective, the use of LLMs aims at reducing the number of experts involved while maintaining the same annotation quality.}

\begin{figure}[!t]
  \centering
  \includegraphics[width=0.46\textwidth, trim=4.3cm 3.8cm 9cm 3.9cm,
  clip]{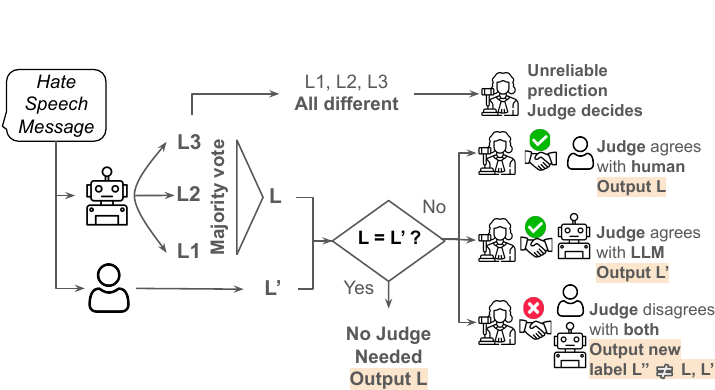}
  \caption{\bo{LLM-in-the-loop framework overview. \dc{A HS message is annotated by a human and an LLM in parallel.} The LLM generates three labels per instance and assigns a majority vote label. If it matches the human label, it is accepted; otherwise, a judge assigns the final label (also when all three LLM labels differ). LLM labels L1–L3; L = LLM majority vote label; L' = human label; L'' = judge label.}}
  \label{fig:pipeline-llm-in-the-loop}
\end{figure}

\tcarr{We illustrate the framework in Figure~\ref{fig:pipeline-llm-in-the-loop}.} In the first stage, a human annotator and an LLM annotate the entire dataset. \dc{When the LLM annotations are not consistent or when the LLM and the human disagree, the claims are assigned to a judge annotator.} 
\dc{While the judge can observe the assigned labels, they can disagree with both and assign a different label. The judge's decision is final.}
\bo{To mitigate anchor bias, the judge is blind to the source of each annotation and receives the labels in a randomly swapped order.} Although both humans and LLMs interact in our setting, \tcarr{our framework differs from a} 
human-in-the-loop approach. In the latter, an automated system completes the task end-to-end, and a human subsequently verifies the output to ensure correctness, generally obtaining silver-quality annotations. In contrast, in the LLM-in-the-loop framework, 
the LLM is integrated into the workflow \bo{as an \textit{in-vitro}, auxiliary annotator, operating in a controlled, human-supervised setting, \tcarr{reducing the costs in terms of manual effort and time.}} 
\tcarr{To mitigate known limitations of using LLM-as-annotators~\citep{baumann2025largelanguagemodelhacking}, we have implemented the following strategies:} \bo{(i.) generate three independent annotations per instance and assign a provisional label via majority voting, reporting prediction variability; (ii.) require human judge when the three LLM annotations disagree or when the LLM majority-vote label conflicts with the first human annotator’s label; (iii.) compare zero-shot and one-shot settings by measuring reductions in prediction variability; (iv.) evaluate judge preferences under blind conditions regarding the annotation source; (v.) validate each step by measuring percent agreement and IAA between LLM annotations, other LLMs, and human annotations; and (vi.) validate the overall framework by comparing LLM-in-the-loop annotations with a fully human annotation setting.} 

\section{Data}
\label{sec:wsf-arg-plus}

For our study, we use the White Supremacy Forum Argumentative dataset (WSF-ARG)~\citep{bonaldi-etal-2024-safer}. 
WSF-ARG comprises 227 hateful messages 
specifically selected as the longest argumentative messages containing at least one premise and one conclusion. While individual argument components may be hateful or not, each message is classified as hateful overall.

\tc{We selected WSF-ARG 
because it contains both explicit and implicit hate speech from real users with sufficient complexity to provide claims suitable for fact-checking assessment. 
The dataset offers valuable features, including identified argument components, component-level hate speech annotations (with some messages classified as hateful despite containing no hateful components), and identification of the weakest argument component in each message.} \tcarr{However,} \bo{WSF-ARG does not contain \tcarr{non-}hateful messages.} \tcarr{Since we aim at investigating the relationship between check-worthiness and hate speech, we have extended WSF-ARG} 
with 294 non-hateful messages. \tcarr{This allows us to investigate whether check-worthiness supports the identification of hate speech.} 

To do this extension, we followed the approach of \citet{bonaldi-etal-2024-safer}. \tcarr{First, } we selected the longest non-hateful messages from the original WSF dataset (exceeding 40 words) and applied \texttt{gpt-3.5-turbo-instruct} to automatically extract claims. \tcarr{After this,} two human annotators\footnote{Sociodemographic of the annotators are in Appendix~\ref{sec:annotators-demographics}.} 
validated the extracted claims by comparing them with the original messages. Messages with no claims, premises, or conclusions were discarded 
\tcarr{If a message was deemed as valid, but the extraction of claims was imperfect, the annotators corrected it with minimal edits. Messages requiring substantial rewriting were discarded.} 

\bo{Table \ref{tab:wsf-stats} presents \tcarr{an overview of the enriched WSF-ARG. As the table shows, hateful messages} 
contain at least two claims, one conclusion (always present on hateful messages in WSF-ARG) and one or multiple premises that support the conclusion (2.789 claims on avg. per hateful message). In contrast, non-hateful messages do not consistently follow a premise–conclusion structure (only 136 out of 294 contain an explicit conclusion), although they still consist of one or multiple claims (with an average of 3.344 claims per message).}


\begin{table}[t]
\resizebox{!}{2cm}{\
\bo{
\begin{tabular}{lcc}
\toprule
                                     & \multicolumn{1}{l}{HS} & \multicolumn{1}{l}{Non-HS} \\ \midrule
Number of Messages                   & 227                             & 294                                 \\
Total Number of Premises             & 406                             & 362                                 \\
Total Number of Conclusions          & 227                             & 136                                 \\
Total Number of Claims  & 633                             & 982                                 \\
Avg. Number of Claims per Message & 2.789                          & 3.344                              \\
Std. Number of Claims per Message & 0.644                          & 1.160                              \\
Min Number of Claims per Message  & 2                               & 1                                   \\
Max Number of Claims per Message  & 4                               & 7                                  \\ \bottomrule
\end{tabular}}
}
\caption{Descriptive statistics of claims per message in WSF-ARG, enriched with Non-HS messages.}
\label{tab:wsf-stats}
\end{table}

\section{LLM-in-the-loop Evaluation}
\label{sec:experimental-settings}


We use 12 models belonging to five different families, namely Mistral, Llama3*, Qwen2.5, Olmo2, and Commander-R. Models vary in sizes, ranging between 7B (small models) up to 104B (large models).\footnote{\bo{The specific versions of the LLMs used in this study are provided in Appendix~\ref{sec:appendix-llms-version}, along with the energy consumed by these LLMs in our experiments in Appendix~\ref{sec:energy-llms}.}} \tc{We selected these models to ensure that each has a counterpart within the same family but with different parameter sizes. This design enables systematic exploration of how model scale and model family influence the effectiveness of our mitigation strategies. Specifically, this selection allows us to: compare output variability across different LLMs, evaluate multiple model configurations rather than relying on a single LLM, and examine inter-model agreement patterns.} \bo{We have excluded closed models due to replicability concerns.}


\tcarr{Regarding temperature, values in the 0–1 range show no statistically significant differences in performance, while values above 1 degrade it~\citep{renze-2024-effect}. We set temperature to 1 to balance stability and stochasticity for measuring intra-model reliability, with gains expected to generalize to lower settings~\citep{renze-2024-effect,LI2025242}.}

Two expert human annotators 
participate in the task, with one annotating the entire dataset alongside the LLMs and the other serving as the judge. Human annotators and LLMs are given the same instructions and label definitions. All model annotations are generated using a shared base prompt (see prompts in Appendix~\ref{sec:prompts-cw}). 


We further evaluate one-shot prompting, using as examples the messages used to define the labels in Section~\ref{sec:cw-detection}, \boarxiv{hypothesizing that one-shot prompting may reduce LLM prediction variability before the majority voting stage and avoiding inconsistent labeling.} 

Given that we use a dataset in which claims are explicitly extracted from each message, we annotate check-worthiness per claim instead of per message. This allows for a more controlled and nuanced analysis of our experiments. The proposed LLM-in-the-loop framework can also be applied to entire messages where claim level annotations are not available.

\subsection{Results \& Discussion}
\label{sec:results}

\begin{table*}[t]
\centering
\resizebox{!}{4.8cm}{\
\begin{tabular}{
 l  
 l |
 c 
c 
c 
c |
c 
c 
c 
c }
\toprule                                    
                                  &        &        \multicolumn{4}{c|}{Hate Speech Messages}                    & \multicolumn{4}{c}{Non-Hate Speech Messages} \\ \midrule
\multicolumn{1}{c}{Model}         & Shot & All Claims               & CFS                      & NFS                         & UFS                              & All Claims          & CFS                & NFS                & UFS     \\ \midrule
                                  & zero   & .346             $\pm$ .001  & .096            $\pm$ .000  &    \underline{.994         $\pm$ .000}     & .196            $\pm$ .006         & .465            $\pm$ .003                & .365            $\pm$ .003                   & .886             $\pm$ .002             & .256            $\pm$ .006  \\ 
\multirow{-2}{*}{Mistral-7B}      & one    & .408             $\pm$ .002  & .219            $\pm$ .004  &    .806         $\pm$ .003     & \textbf{.441    $\pm$ .006}        & .526            $\pm$ .001                & .378            $\pm$ .000                   & .670             $\pm$ .002             & \textbf{.570    $\pm$ .002}  \\ 
                                  & zero   & .292             $\pm$ .003  & .027            $\pm$ .003  &    .988         $\pm$ .003     & .118            $\pm$ .011         & .383            $\pm$ .002                & .119            $\pm$ .006                   & .951             $\pm$ .004             & .230            $\pm$ .004  \\ 
\multirow{-2}{*}{Llama-8B}        & one    & .324             $\pm$ .001  & .082            $\pm$ .002  &    .964         $\pm$ .000     & .157            $\pm$ .000         & .435            $\pm$ .005                & .227            $\pm$ .003                   & .966             $\pm$ .000             & .253            $\pm$ .011  \\ 
                                  & zero   & .389             $\pm$ .008  & .213            $\pm$ .014  &    .855         $\pm$ .014     & .265            $\pm$ .020         & .477            $\pm$ .002                & .494            $\pm$ .008                   & .758             $\pm$ .017             & .250            $\pm$ .008  \\ 
\multirow{-2}{*}{Olmo2-7B}        & one    & .322             $\pm$ .008  & .090            $\pm$ .004  &    .964         $\pm$ .013     & .118            $\pm$ .025         & .425            $\pm$ .003                & .304            $\pm$ .009                   & .905             $\pm$ .010             & .191            $\pm$ .003  \\ 
                                  & zero   & .363             $\pm$ .002  & .104            $\pm$ .003  &    .964         $\pm$ .000     & .324            $\pm$ .011         & .455            $\pm$ .002                & .229            $\pm$ .003                   & .932             $\pm$ .004             & .331            $\pm$ .003  \\ 
\multirow{-2}{*}{Qwen2.5-7B}      & one    & .311             $\pm$ .002  & .055            $\pm$ .002  &    .994         $\pm$ .003     & .127            $\pm$ .011         & .364            $\pm$ .001                & .124            $\pm$ .000                   & \underline{.996             $\pm$ .000}             & .138            $\pm$ .002  \\ 
                                  & zero   & .458             $\pm$ .007  & .418            $\pm$ .013  &    .800         $\pm$ .012     & .049            $\pm$ .020         & .408            $\pm$ .005                & .467            $\pm$ .003                   & .814             $\pm$ .011             & .048            $\pm$ .007  \\ 
\multirow{-2}{*}{Command-R-7B}    & one    & .423             $\pm$ .007  & .339            $\pm$ .010  &    .855         $\pm$ .003     & .029            $\pm$ .000         & .408            $\pm$ .005                & .439            $\pm$ .013                   & .856             $\pm$ .004             & .045            $\pm$ .006  \\ \midrule
                                  & zero   & \underline{.589  $\pm$ .003} & \textbf{.697    $\pm$ .004} &    .588         $\pm$ .010     & .206            $\pm$ .006         & .518            $\pm$ .007                & \underline{.657 $\pm$ .016}                  & .652             $\pm$ .015             & .278            $\pm$ .017  \\ 
\multirow{-2}{*}{Mixtral-8x7B}    & one    & .585             $\pm$ .007  & \underline{.661 $\pm$ .007} &    .533         $\pm$ .009     & \underline{.392 $\pm$ .034}        & .550            $\pm$ .011                & .569            $\pm$ .004                   & .557             $\pm$ .023             & \underline{.525 $\pm$ .037}  \\
                                  & zero   & .336             $\pm$ .003  & .123            $\pm$ .003  &    .988         $\pm$ .003     & .049            $\pm$ .006         & .387            $\pm$ .001                & .298            $\pm$ .003                   & .992             $\pm$ .000             & .028            $\pm$ .000  \\ 
\multirow{-2}{*}{Mistral-22B}     & one    & .303             $\pm$ .000  & .063            $\pm$ .000  &    \textbf{1.0  $\pm$ .000}    & .039            $\pm$ .000         & .325            $\pm$ .001                & .141            $\pm$ .002                   & \textbf{1.0      $\pm$ .004}            & .011            $\pm$ .002  \\
                                  & zero   & \textbf{.594     $\pm$ .008} & .620            $\pm$ .004  &    .745         $\pm$ .036     & .255            $\pm$ .006         & \textbf{.581 $\pm$ .003}               & \textbf{.699    $\pm$ .006}                  & .826             $\pm$ .002             & .281            $\pm$ .003  \\ 
\multirow{-2}{*}{Olmo2-32B}       & one    & .507             $\pm$ .001  & .434            $\pm$ .003  &    .927         $\pm$ .003     & .088            $\pm$ .006         & .489            $\pm$ .005                & .508            $\pm$ .013                   & .966             $\pm$ .004             & .115            $\pm$ .002  \\ 
                                  & zero   & .466             $\pm$ .004  & .377            $\pm$ .006  &    .927         $\pm$ .007     & .039            $\pm$ .015         & .488            $\pm$ .008                & .580            $\pm$ .008                   & .947             $\pm$ .010             & .053            $\pm$ .013  \\ 
\multirow{-2}{*}{Mixtral-8x22B}   & one    & .510             $\pm$ .001  & .432            $\pm$ .002  &    .927         $\pm$ .006     & .118            $\pm$ .010         & .536            $\pm$ .003                & .597            $\pm$ .002                   & .951             $\pm$ .004             & .166            $\pm$ .010  \\ \midrule
                                  & zero   & .479             $\pm$ .002  & .333            $\pm$ .003  &    .988         $\pm$ .000     & .176            $\pm$ .015         & .540            $\pm$ .006                & .539            $\pm$ .004                   & .985             $\pm$ .002             & .211            $\pm$ .013  \\ 
\multirow{-2}{*}{Llama-70B}       & one    & .455             $\pm$ .010  & .284            $\pm$ .009  &    .988         $\pm$ .000     & .206            $\pm$ .030         & .497            $\pm$ .003                & .478            $\pm$ .003                   & \underline{.996             $\pm$ .000}             & .146            $\pm$ .009  \\ 
                                  & zero   & .485             $\pm$ .002  & .290            $\pm$ .002  &    .976         $\pm$ .003     & \underline{.392            $\pm$ .010}         & \underline{.569            $\pm$ .004}                & .442            $\pm$ .006                   & .981             $\pm$ .006             & .393            $\pm$ .010  \\ 
\multirow{-2}{*}{Qwen2.5-72B}     & one    & .453             $\pm$ .003  & .246            $\pm$ .002  &    .988         $\pm$ .000     & .333            $\pm$ .026         & \underline{.569    $\pm$ .003}               & .428            $\pm$ .004                   & .989             $\pm$ .002             & .402            $\pm$ .006  \\ 
                                  & zero   & .545             $\pm$ .004  & .549            $\pm$ .010  &    .867         $\pm$ .009     & .010            $\pm$ .010         & .453            $\pm$ .027                & .511            $\pm$ .018                   & .951             $\pm$ .096             & .025            $\pm$ .016  \\ 
\multirow{-2}{*}{Command-R-104B}  & one    & .321             $\pm$ .004  & .101            $\pm$ .007  &    \textbf{1.00         $\pm$ .000}     & .010            $\pm$ .006         & .345            $\pm$ .004                & .188            $\pm$ .014                   & \underline{.996  $\pm$ .000}            & .022            $\pm$ .011  \\ \bottomrule
\end{tabular}}
\caption{\bo{Percentage agreement between the first human annotator and LLM predictions across 24 configurations. Agreement is calculated as matching labels divided by total instances. Values show percent agreement of majority-vote LLM annotations with the first annotator (± standard deviation across three runs). The highest value per column is in bold and the second highest is underlined.}}
\label{tab:agreement}
\end{table*}

\tcarr{We begin by identifying the best-performing model and prompt configuration for check-worthiness annotation. Table~\ref{tab:agreement} reports agreement between each of the 24 configurations, spanning 12 models and two prompt strategies—and the first human annotator.} OLMo2-32B attains the highest agreement, with a percentage agreement of 0.594 on claims from hate speech messages and 0.581 on claims from non-hate speech messages, followed by Mixtral-8x7B with corresponding scores of 0.589 and 0.518.
Overall, both zero-shot and one-shot approaches exhibit minimal variability. Larger models do not exhibit higher agreement with human annotators compared to small or medium-sized ones. Models show stronger agreement with the first annotator on identifying CFS labels, while disagreement is more frequent for NFS and UFS labels. 

\tcarr{OLMo2-32B achieves moderate agreement with the first annotator on check-worthiness annotation (Cohen's $\kappa$ = 0.365), reaching Cohen's $\kappa$ = 0.348 on claims contained in hate speech messages, and Cohen's $\kappa$ = 0.350 on claims contained in non-hate speech messages. Agreement improves under the binary setting (merging NFS and UFS into a single Non-Check-Worthy label), reaching Cohen's $\kappa$ = 0.376 on claims contained in hate speech messages and Cohen's $\kappa$ = 0.516 on claims contained in non-hate speech messages, with an overall Cohen's $\kappa$ of 0.483, comparable to human agreement reported in the 2021 CheckThat! Lab~\citep{shaar2021overview}, 
and with non-hate speech 
consistently outperforming hate speech, suggesting that check-worthiness annotation is inherently harder for hateful content.} 

\begin{table}[t]
\centering
\resizebox{!}{2.38cm}{
\begin{tabular}{l|ccc|ccc}
\toprule
                         & \multicolumn{3}{c|}{Zero Shot ($\kappa$ Score)}                                                                       & \multicolumn{3}{c}{One Shot ($\kappa$ Score)}                                                                        \\
                         & \multicolumn{1}{c}{Non-HS} & \multicolumn{1}{c}{HS} & \multicolumn{1}{c|}{Overall} & \multicolumn{1}{c}{Non-HS} & \multicolumn{1}{c}{HS} & \multicolumn{1}{c}{Overall} \\ \midrule
Mistral-7B       & 0.387  & 0.147   & 0.267   & 0.419   & 0.244  & 0.332 \\
Llama-8B         & 0.303  & 0.126   & 0.214   & 0.361   & 0.160  & 0.261 \\
Olmo2-7B         & 0.448  & 0.291   & 0.370   & 0.314   & 0.127  & 0.220 \\
Qwen2.5-7B       & 0.386  & 0.172   & 0.279   & 0.285   & 0.123  & 0.204 \\
Command-r-7B     & 0.377  & 0.209   & 0.293   & 0.349   & 0.207  & 0.278 \\ \midrule
Mixtral-8x7B     & 0.476  & 0.418   & 0.447   & 0.457   & 0.453  & 0.455 \\
Mistral-22B      & 0.322  & 0.150   & 0.236   & 0.160   & 0.065  & 0.112 \\
Olmo2-32B        & 1.000  & 1.000   & 1.000   & 0.562   & 0.541  & 0.551 \\
Mixtral-8x22B    & 0.551  & 0.409   & 0.480   & 0.563   & 0.475  & 0.519 \\ \midrule
Llama-70B        & 0.553  & 0.405   & 0.479   & 0.500   & 0.337  & 0.419 \\
Qwen2.5-72B      & 0.555  & 0.391   & 0.473   & 0.518   & 0.329  & 0.423 \\
Command-R-104B   & 0.511  & 0.464   & 0.487   & 0.232   & 0.112  & 0.172 \\ 
\bottomrule
\end{tabular}}
\caption{IAA (Cohen's $\kappa$) of the best-performing configuration, \textbf{Olmo2-32B} under zero-shot with majority voting, compared against other LLM runs (also using majority voting).}
\label{tab:cohen-agreement}
\end{table}

\paragraph{Prediction variability} 
LLMs exhibit high consistency when predicting claim labels: in the worst-case scenario, Command-R-7B (zero-shot, hate speech) and Command-R-104B (zero-shot, non-hate speech) produce identical labels in all three runs for 77.7\% and 70.7\% of cases, respectively. In most remaining instances, two out of three runs agree. Only Mixtral-8x7B in the zero-shot setting produces different labels across all three runs, occurring in fewer than 3\% of cases. \tcarr{One-shot prompting generally yields more stable labels across repeated runs of the same LLM. It rarely produces three different labels across three runs, indicating improved robustness even at a temperature of 1, albeit with a possible reduction in overall agreement. Full details in Table~\ref{tab:variability} in Appendix~\ref{sec:pred-var}.}


\paragraph{Annotator Agreement} \tcarr{OLMo2-32B achieves the highest agreement with the first human annotator under zero-shot prompting and shows low variability. Table~\ref{tab:cohen-agreement} further shows moderate agreement between OLMo2-32B and other medium- and large-scale models, confirming consistent behavior across configurations and supporting its selection 
to be judged in the second phase.
754 of the total claims are required to be judged (46.69\%), these claims appeared more frequently in Non-HS messages (497; 50.61\% of claims in Non-HS) than in HS messages (257; 40.60\% of claims in HS). 
The judge sides with the human annotator in 589 claims (78.12\% of claims to be judged) and with the LLM in 138 claims (18.30\% of claims to be judged). 
After adjudication, disagreements decrease from 757 (46.69\%) to 616 claims (38.14\%)}. \bocr{This reduction occurs because, in a subset of disputed cases, the judge agrees with the LLM, bringing those LLM predictions into agreement. The remaining disagreement therefore reflects the divergence between OLMo2-32B's zero-shot predictions and the final adjudicated labels produced by the LLM-in-the-loop framework.} 

\subsection{Human Validation}
\label{sec:human-val}
\bocr{To verify whether the labels produced by the LLM-in-the-loop framework align with those obtained through full human annotation, three expert annotators independently annotated the entire dataset using the same annotation guidelines. These annotations consisted of the re-used annotations from the first annotator involved in the LLM-in-the-loop setup (without relabeling) and new annotations provided by two additional expert annotators.
In cases of disagreement among the three annotators, a fourth human annotator (who was not involved in the LLM-in-the-loop setup) reviewed the instance. Otherwise, the final label was determined through majority voting.
Note that, with this setup, no memory effects are introduced into the annotation process. All annotators involved in both the LLM-in-the-loop and full-human setups 
saw their assigned claims only once, avoiding the risk of recency effects.}

The IAA agreement between pairs of annotators using Cohen's $\kappa$ ranges between 0.485 and 0.627, suggesting moderate to substantial agreement (see details in Appendix~\ref{sec:iaa-all}). Using Krippendorff's $\alpha$ across all annotators yields 0.537 when considering the three labels (NFS, UFS, and CFS), and 0.570 when using a binary formulation in which NFS and UFS are grouped as Non Check-worthy and CFS as Check-worthy. For this setting, only 73 cases required a judge (4.52\% of the total claims), followed by 699 cases where two annotators agreed, and one disagreed (43.28\% of the total claims), and 843 cases where all annotators agreed (52.20\% of the total claims). From the 73 claims to be judged in the human setting, 53 (72.6\%) matched cases to be judged in the LLM-in-the-loop setting.
\bo{We then compared the \tcarr{final} check-worthiness annotations obtained through the LLM-in-the-loop framework with those produced through full human annotation. IAA was calculated, yielding a Cohen's $\kappa$ of 0.800 when considering the three labels (NFS, UFS, and CFS), and 0.815 when using a binary formulation \tcarr{with NFS and UFS are grouped as Non Check-worthy.} 
Table \ref{tab:label-dist} presents the distribution of check-worthiness annotations across HS and Non-HS messages for both LLM-in-the-loop and full human annotation methods, further broken down by claims, showing that the annotations of both settings distribute similarly even when intersected with HS.}
\bo{With this human validation, we can see that, although LLM-as-annotators may produce unreliable labels (as highlighted by \citet{baumann2025largelanguagemodelhacking}), our LLM-in-the-loop framework can mitigate their impact even for a highly subjective task such as check-worthiness detection.} 

\section{WSF-ARG+: A New Dataset for Hate Speech and Check-worthiness Detection}

\begin{table}[t]
\centering
\resizebox{!}{2.7cm}{
\begin{tabular}{lrrrrrr}
\toprule
& \multicolumn{6}{c}{LLM-in-the-loop (gold labels)} \\
   & \multicolumn{3}{l}{Claims on HS Mess.} & \multicolumn{3}{l}{Claims on Non-HS Mess.}                         \\ \midrule
HS Label \\ per Claim & CFS          & NSF          & UFS         & CFS                   & NFS                   & UFS                   \\ \midrule
Non-HS                      & 132          & 75           & 60          &                       &                       &                       \\
HS                          & 191          & 142          & 33          & \multirow{-2}{*}{348} & \multirow{-2}{*}{318} & \multirow{-2}{*}{316} \\ 
\midrule
& \multicolumn{6}{c}{Full Human (platinum labels)} \\
& \multicolumn{3}{l}{Claims on HS Mess.} & \multicolumn{3}{l}{Claims on Non-HS Mess.}                         \\ \midrule
HS Label \\ per Claim & CFS          & NSF          & UFS         & CFS                   & NFS                   & UFS                   \\ \midrule
Non-HS                      & 131          & 59           & 77          &                       &                       &                       \\
HS                          & 183          & 143          & 40          & \multirow{-2}{*}{327} & \multirow{-2}{*}{339} & \multirow{-2}{*}{316} \\ 
\bottomrule
\end{tabular}}
\caption{
Label distributions for CFS, UFS, and NFS across HS and Non-HS messages and their constituent claims in the WSF-ARG dataset, comparing the full human setting with the LLM-in-the-loop framework using \textbf{OLMo2-32B} (zero-shot, majority voting).}
\label{tab:label-dist}
\end{table}

\bo{As a result of our LLM-in-the-loop framework and extensive validation, we introduce WSF-ARG+, a new dataset extending WSF-ARG. \tcarr{Next to the 294 non-hateful messages, it also contains annotations for check-worthiness at claim level for each message.} 
For each claim, we release all annotations produced by the LLM-in-the-loop framework \tcarr{and humans}, both aggregated and disaggregated. \tcarr{We refer to the full human annotations as \textit{platinum labels}, while those from the LLM-in-the-loop are called \textit{gold labels}.} 
We also release the predictions of all LLMs. 
WSF-ARG+ establishes 
the first testbed for both detection and counterspeech generation at the intersection of hate speech and check-worthiness.} 
\bo{Table \ref{tab:label-dist} reports on the label distribution of WSF-ARG+. We can notice that, for this dataset, hateful messages rely more heavily on check-worthy claims to convey hate. These claims may themselves be hateful or not. In contrast, although check-worthy claims also appear in non-hateful messages, such messages predominantly do not depend on check-worthy claims.}


\section{Advancements at the Intersection of Hate Speech and Check-Worthiness}
\label{sec:results-hs-det-and-openaitool}

We further analyze whether check-worthy hateful messages are more harmful than not check-worthy ones. \bocr{To this end, we rely on the OpenAI Moderation Tool\footnote{\url{https://platform.openai.com/docs/guides/moderation}} as it achieves good results for hate speech detection, reaching the best performance on HateModerate \cite{zheng-etal-2024-hatemoderate}, a benchmark dataset built from several well-known datasets in the hate speech domain, such as HateCheck, DynaHate, ToxicSpans, Hate Offensive, to mention a few. The OpenAI moderation tool outputs, for an input message, 21 dimension scores ranging from 0 to 1, where higher values indicate a higher likelihood of the content containing aspects of that dimension. However, OpenAI does not provide a unified threshold either within or across dimensions. In our experiments, we focus on \textit{hate} and \textit{harassment}, as they are the only ones exhibiting average scores above 0.1 for messages in WSF-ARG+. Dimensions with lower scores are excluded since they do not meaningfully trigger the corresponding categories.} 

\bocr{To complement the validation conducted by \citet{zheng-etal-2024-hatemoderate}, we evaluated the entire WSF-ARG+ dataset using the OpenAI Moderation tool, targeting the hate and harassment dimensions, and compared its outputs against the ground-truth hate speech labels (227 HS and 294 Non-HS). Table \ref{tab:openai-validation} reports their macro precision (P), recall (R), and F1 scores for thresholds ranging from 0.1 to 0.9. At a threshold of 0.1, the tool achieves macro F1 of 0.808 (hate) and 0.764 (harassment), well above a random baseline of 0.5, demonstrating reliable alignment with ground-truth labels. Scores remain consistently above the random baseline across all thresholds for harassment, and for hate up to threshold 0.8. 
We note that while the tool's internal policies may not perfectly mirror the definition of hate speech in WSF-ARG+ (and consequently the actual annotations), there is a strong empirical alignment.}

\begin{table}[t]
\small
\begin{tabular}{l|rrr|rrr}
\toprule
th & $\text{P}_{\text{hate}}$ & $\text{R}_{\text{hate}}$ & $\text{F}1_{\text{hate}}$ & $\text{P}_{\text{harass}}$ & $\text{R}_{\text{harass}}$ & $\text{F}1_{\text{harass}}$ \\
\midrule
0.1 & .822 & .803 & \textbf{.808} & .789 & .782 & .764 \\
0.2 & .824 & .765 & .770 & .799 & .804 & \textbf{.798} \\
0.3 & .824 & .765 & .770 & .794 & .798 & .795 \\
0.4 & .806 & .700 & .695 & .797 & .798 & .797 \\
0.5 & .795 & .673 & .661 & .800 & .797 & \textbf{.798} \\
0.6 & .795 & .643 & .618 & .802 & .794 & .797 \\
0.7 & .792 & .623 & .587 & .774 & .749 & .753 \\
0.8 & .806 & .588 & .529 & .777 & .745 & .749 \\
0.9 & .787 & .518 & .398 & .780 & .654 & .636 \\
\bottomrule
\end{tabular}
\caption{\bocr{Macro precision (P), recall (R), and F1 scores for the \textit{hate} (hate) and \textit{harassment} (harass) dimensions across thresholds (th) ranging from 0.1 to 0.9 on WSF-ARG+.}} \label{tab:openai-validation}
\end{table}

\begin{table}[t]
\centering
\resizebox{!}{1.45cm}{\
\begin{tabular}{l|rr|rr}
\toprule
 & \multicolumn{2}{c|}{Platinum} & \multicolumn{2}{c}{Gold} \\
                                                    & Harass         & Hate             & Harass & Hate      \\ \midrule
HS Mess. w/o CFS claims                             & 0.611          & 0.319            & 0.638 & 0.317      \\
HS Mess. w/ $\geq 1$ CFS claim                  & \textbf{0.728*} & \textbf{0.411*}   & \textbf{0.713*} & \textbf{0.405}      \\ \midrule
Stat                                                & 3872.5         & 4236.5           & 3479.5 & 3723.0    \\ 
P-value                                             & \textbf{0.005}          & \textbf{0.038}            & \textbf{0.022} & 0.078      \\
Effect Size                                         & 0.227          & 0.154            & 0.190 & 0.133      \\ \bottomrule
\end{tabular}}
\caption{Comparison of OpenAI moderation scores for harassment and hate between HS messages w/ and w/o Check-worthy claims, including Mann–Whitney U statistics, p-values, and effect sizes; * denotes statistically significance ($\text{p-value} < 0.05$).} \label{tab:openai-scores}
\end{table}

\begin{table*}[t]
\centering
\resizebox{!}{2.89cm}{\
\begin{tabular}{l|cccc|cccc|ccc}
\toprule
\multirow{2}{*}{Models} & \multicolumn{4}{c|}{w/ check-worthiness (platinum)} & \multicolumn{4}{c|}{w/ check-worthiness (gold)} & \multicolumn{3}{c}{w/o check-worthiness} \\
                & P               & R               & F1                       & $\Delta$ F1               & P               & R               & F1                        & $\Delta$ F1    & P             & R             & F1                              \\ \midrule
Mistral-7B	    & .648 $\pm$ .002 & .611 $\pm$ .002 & .560 $\pm$ .003          & -.117          & .677 $\pm$ .000 & .662 $\pm$ .000 & .636          $\pm$ .000  & -.041          & .738 $\pm$ .000 & 0.709  $\pm$ .000 & \textbf{.677 $\pm$ .000}  \\ 
Llama-8B	    & .711 $\pm$ .002 & .564 $\pm$ .003 & .438 $\pm$ .005          & -.034          & .644 $\pm$ .000 & .568 $\pm$ .000 & .472          $\pm$ .000  & .000           & .624 $\pm$ .000 & 0.563  $\pm$ .000 & \textbf{.472 $\pm$ .000}  \\ 
Olmo2-7B	    & .746 $\pm$ .005 & .746 $\pm$ .005 & \textbf{.746 $\pm$ .005} & \textbf{+.059*}  & .782 $\pm$ .004 & .786 $\pm$ .004 & \textbf{.776  $\pm$ .004} & \textbf{+.089*} & .768 $\pm$ .001 & 0.724  $\pm$ .001 & .687          $\pm$ .001  \\ 
Qwen2.5-7B	    & .759 $\pm$ .000 & .639 $\pm$ .000 & \textbf{.559 $\pm$ .000} & \textbf{+.031}  & .734 $\pm$ .005 & .673 $\pm$ .003 & \textbf{.621  $\pm$ .002} & \textbf{+.093*} & .745 $\pm$ .000 & 0.619  $\pm$ .000 & .528          $\pm$ .000  \\ 
Command-r-7B	& .497 $\pm$ .000 & .498 $\pm$ .000 & .426 $\pm$ .000          & -.229          & .557 $\pm$ .000 & .523 $\pm$ .000 & .409          $\pm$ .000  & -.246          & .729 $\pm$ .000 & 0.692  $\pm$ .000 & \textbf{.655  $\pm$ .000} \\ \midrule
Avg Small	    & .672 $\pm$ .002 & .612 $\pm$ .002 & .546 $\pm$ .003          & -.058          & .679 $\pm$ .002 & .642 $\pm$ .001 & .583          $\pm$ .001  & -.021          & .721 $\pm$ .000 & 0.661  $\pm$ .000 & \textbf{.604  $\pm$ .000} \\ \midrule
Mixtral-8x7B	& .621 $\pm$ .005 & .589 $\pm$ .004 & .536 $\pm$ .005          & -.113          & .613 $\pm$ .008 & .570 $\pm$ .009 & .498          $\pm$ .015  & -.151          & .730 $\pm$ .007 & 0.688  $\pm$ .009 & \textbf{.649  $\pm$ .012} \\ 
Mistral-22B	    & .596 $\pm$ .001 & .568 $\pm$ .002 & .510 $\pm$ .005          & -.125          & .648 $\pm$ .005 & .607 $\pm$ .005 & .553          $\pm$ .006  & -.082          & .712 $\pm$ .002 & 0.675  $\pm$ .001 & \textbf{.635  $\pm$ .001} \\ 
Olmo2-32B	    & .722 $\pm$ .020 & .725 $\pm$ .020 & .722 $\pm$ .020          & -.057          & .772 $\pm$ .000 & .774 $\pm$ .000 & .773          $\pm$ .000  & -.006          & .787 $\pm$ .000 & 0.790  $\pm$ .000 & \textbf{.779  $\pm$ .000} \\ 
Mixtral-8x22B	& .754 $\pm$ .005 & .749 $\pm$ .005 & .751 $\pm$ .005          & -.034          & .757 $\pm$ .004 & .760 $\pm$ .003 & .758          $\pm$ .004  & -.027          & .786 $\pm$ .003 & 0.790  $\pm$ .003 & \textbf{.785  $\pm$ .003} \\ \midrule
Avg Medium	    & .673 $\pm$ .008 & .658 $\pm$ .008 & .630 $\pm$ .009          & -.082          & .698 $\pm$ .004 & .678 $\pm$ .004 & .645          $\pm$ .006  & -.067          & .754 $\pm$ .003 & 0.736  $\pm$ .003 & \textbf{.712  $\pm$ .004} \\ \midrule
Llama-70B	    & .686 $\pm$ .011 & .679 $\pm$ .014 & \textbf{.659 $\pm$ .017} &  \textbf{+.179*} & .704 $\pm$ .005 & .688 $\pm$ .007 & \textbf{.662  $\pm$ .009} & \textbf{+.182*} & .675 $\pm$ .008 & 0.579  $\pm$ .006 & .480          $\pm$ .009  \\ 
Qwen2.5-72B	    & .723 $\pm$ .004 & .725 $\pm$ .004 & \textbf{.716 $\pm$ .003} &  \textbf{+.015*} & .773 $\pm$ .006 & .775 $\pm$ .006 & \textbf{.766  $\pm$ .007} & \textbf{+.065*} & .766 $\pm$ .014 & 0.733  $\pm$ .016 & .701          $\pm$ .018  \\ 
Command-R-104B	& .698 $\pm$ .003 & .623 $\pm$ .004 & \textbf{.553 $\pm$ .006} &  \textbf{+.173*} & .713 $\pm$ .003 & .650 $\pm$ .005 & \textbf{.593  $\pm$ .007} & \textbf{+.213*} & .705 $\pm$ .001 & 0.535  $\pm$ .001 & .380          $\pm$ .002  \\ \midrule
Avg Large	    & .702 $\pm$ .006 & .676 $\pm$ .007 & \textbf{.643 $\pm$ .009} &  \textbf{+.123} & .730 $\pm$ .005 & .704 $\pm$ .006 & \textbf{.674  $\pm$ .008} & \textbf{+.154} & .715 $\pm$ .008 & 0.615  $\pm$ .008 & .520          $\pm$ .009  \\ \bottomrule 
\end{tabular}} \caption{Macro precision (P), recall (R), and F1 results for binary hate speech detection on WSF-ARG+. Scores are averaged over three runs; ($\pm$) denotes the standard deviation. $\Delta$F1 indicates the change in macro F1 when check-worthiness is included versus excluded. Positive values indicate improvement, while negative values indicate a decline in performance; * denotes a statistically significant difference according to McNemar's test ($\text{p-value} < 0.05$).}
\label{tab:hs-clf-scores-with-std}
\end{table*}

\bocr{Building on this validation}, we consider two groups: messages that contain at least one check-worthy claim and messages that contain no check-worthy claims. \bocr{We consider both platinum and gold annotations}. 
\tc{Table~\ref{tab:openai-scores} reports the scores for the hate and harassment dimensions, together with \bocr{Mann–Whitney U statistical significance tests} and effect sizes. Differences in harassment and hate are both statistically significant ($\text{p-value} < 0.05$) \bocr{with platinum labels}, indicating that HS messages containing check-worthy claims are, on average, more harassing and more hateful in WSF-ARG+. However, effect sizes are small (harassment: $r=0.227$; hate: $r=0.154$), indicating that while the trend is reliable, the practical distinction between the two groups remains modest. \bocr{Using gold labels, the significant difference is observed only with the harassment dimension ($r=0.190$), while a moderate difference can be seen for the hate dimension ($0.05 < \text{p-value} < 0.1$; $r=0.133$).}}

We further conduct a hate speech classification experiment on the whole WSF-ARG+ dataset. We prompt all 12 LLMs presented  
in Section \ref{sec:experimental-settings} \bocr{over three runs per LLM} to perform a classification task given an input message, and evaluate whether providing check-worthiness labels improves hate speech detection. When using these labels, each claim is wrapped with its corresponding check-worthiness label and embedded directly into the prompt (see prompts in Appendix~\ref{sec:add-details-hs-misinfo}). \bocr{To assess statistical significance, we applied McNemar’s test to the aggregated (via majority voting) LLM predictions across three runs against the ground truth. Table \ref{tab:hs-clf-scores-with-std} reports the mean performance of each LLM over these runs, using both platinum and gold labels.}
The results indicate that all large models ($\geq$ 70B) benefit statistically significantly ($\text{p-value} < 0.05$) from the inclusion of platinum and gold check-worthiness labels, achieving macro F1 scores ranging from 0.553 (Command-R-104B) to 0.766 (Qwen2.5-72B). Across large models, these variants outperform their corresponding versions without check-worthiness labels by an average of 0.123 and 0.154 macro F1 points when using platinum and gold labels, respectively. Within the same model family, Qwen2.5-7B also improves over the setting without check-worthiness labels, reaching an F1 score of \bocr{0.559 with platinum and 0.621 with gold labels, although only the improvement with gold labels is statistically significant}. Among the smaller models also Qwen2.5-7B outperforms statistically significantly their counterparts without check-worthiness, achieving F1 scores of 0.746 with platinum and 0.776 with gold. 

\section{Conclusions and Future Work}
\label{sec:merge_conclusion}

\bo{In this paper, we introduce WSF-ARG+, the first dataset created at the intersection of hate speech and check-worthiness detection. The dataset contains both hateful and non-hateful messages from real users from the White Supremacy Forum Stormfront. The messages in WSF-ARG+ are sufficiently rich in content to contain multiple claims that can be interlinked, can be check-worthy, and can be used by hateful users to intensify and express hateful stances. We release WSF-ARG+ with HS and Non-HS messages, their extracted claims, platinum human check-worthiness annotations (disaggregated and aggregated), and gold annotations obtained through our novel LLM-in-the-loop annotation framework.}

\bo{Along with WSF-ARG+, we also present a step-by-step framework for careful annotation design with LLMs, discussing model selection, prompting strategies, intra- and inter-model comparisons, and extensive human validation. Our results show that, by following our LLM-in-the-loop framework, human effort can be reduced for a 
highly subjective task such as check-worthiness detection in WSF-ARG+, without reducing annotation quality.}

\bo{Finally, we show that hate speech delivered through check-worthy claims is significantly more hateful and associated with higher levels of harassment, and providing check-worthiness annotations as input to LLMs can improve hate speech detection performance for \bocr{large models on WSF-ARG+.}}

\bo{Our findings open several promising directions for future research. 
Check-worthy claims in WSF-ARG+ can be fact-checked to reveal the extent to which HS messages rely on false or misleading claims. WSF-ARG+ also allows for 
new counterspeech generation methods, following~\citet{bonaldi-etal-2024-safer, bonaldi2026catchmeragdatasetcontextually, martone2026assistedcounterspeechwritingcrossroads}, by targeting check-worthy or misinformative claims. 
Finally, investigating the generalizability of the LLM-in-the-loop framework to other annotation tasks could represent a significant step toward improving the reliability of LLMs as annotators.}

\section*{Ethical Statements}

This paper includes examples from the existing hate speech resource WSF-ARG that are used to construct WSG-ARG+. The messages contained in these datasets may include harmful or offensive content and do not reflect the views of the authors.

Although our objective is to contribute to efforts that limit the spread of HS on social media, we acknowledge that releasing WSF-ARG+ may carry a risk of misuse. Nevertheless, we believe that effective strategies to counter HS require diverse research directions, including but not limited to detection, explanation generation, and counterspeech generation. Observations in WSF-ARG+ suggest that HS conveyed through misinformation can be perceived as more hateful. It may also contribute to severe real-world consequences, such as the Rwandan genocide and the persecution of the Rohingya people. Studying the interaction between HS and misinformation may therefore support the development of countering strategies. In particular, aspects such as check-worthiness within HS messages may help identify potential intervention points. Our work aims to encourage further research on the intersection of HS and misinformation, an area that remains relatively underexplored despite its societal importance.

We 
acknowledge that the inclusion of HS content may expose researchers and annotators to harmful language and potential emotional distress. To mitigate this risk, annotators were informed in advance about the nature of the material and participated voluntarily.

\dc{We also note that while HS detection is a subjective task, policies based on HS labels may lead to censorship and speech policing. For this reason, we advocate a counter-speech generation approach based on such labeling, rather than censoring or filtering potentially hateful content.}

Finally, the release of WSF-ARG+ follows the precedent of the previously published resources WSF and WSF-ARG. Consistent with these datasets, WSF-ARG+ is released in a properly anonymised form, without the ability to trace or identify real users, and is intended for research purposes only.

\section*{Limitations}

A primary limitation of this work concerns the generalisability of the proposed LLM-in-the-loop framework. Our validation focuses on check-worthiness detection using data from the white supremacist forum Stormfront. Although the framework was evaluated on a highly subjective task within a similarly subjective domain (hate speech), further studies are required to determine whether it can reduce human effort without compromising annotation quality across other tasks and domains. \dc{Moreover, these studies will have to determine the suitability of LLMs to replace annotators who show a more heterogeneous background than those involved in the experiments described in this paper.} Future work should also examine the extent to which LLMs can replace components of the framework or potentially the entire pipeline, discussing the quality loss compared to full human annotation, which relates to ongoing discussions on LLMs as annotators. While our results support the use of LLM-based annotations, rigorous evaluation is necessary to mitigate risks such as LLM hacking \cite{baumann2025largelanguagemodelhacking}. Finally, the observed improvements in hate speech detection using check-worthiness labels and the finding that hateful content containing check-worthy claims is perceived as more harassing should be validated on datasets beyond WSF-ARG+.

\section*{Acknowledgments}
This work was carried out during the tenure of an ERCIM “Alain Bensoussan" Fellowship Programme. This work used the Dutch national e-infrastructure with the support of the SURF Cooperative using grant no. EINF-15500.
\bocr{The authors would also like to thank Greta Damo and Mariana Chaves for their valuable contributions to the annotation and evaluation of the dataset released in this work, Ekaterina Sviridova for her feedback on the annotation guidelines, and Santiago Marro for his feedback during one brainstorming session. The authors are also grateful to the teams and research labs: 
Marianne @ Centre Inria d'Université Côte d'Azur, CEDAR @ Centre Inria de Saclay, Cyber Intelligence @ IIT-CNR, Trust, Security and Privacy @ IIT-CNR, and PHRLT @ UPV for providing opportunities to present earlier iterations of this work in seminars providing valuable feedback.}
\bibliography{custom.bib}

@misc{ember2026_lifecycle_carbon_intensity,
  author       = {{Ember} and {Our World in Data}},
  title        = {Lifecycle carbon intensity of electricity generation -- Ember},
  year         = {2026},
  howpublished = {[dataset]},
  note         = {With major processing by Our World in Data. Ember, "Yearly Electricity Data Europe"; Ember, "Yearly Electricity Data". Retrieved August 27, 2026 (archived on August 27, 2026)},
  url          = {https://archive.ourworldindata.org/20260827-085718/grapher/electricity-mix.html}
}

@misc{martone2026assistedcounterspeechwritingcrossroads,
      title={Assisted Counterspeech Writing at the Crossroads of Hate Speech and Misinformation}, 
      author={Genoveffa Martone and Helena Bonaldi and Marco Guerini},
      year={2026},
      eprint={2605.22435},
      archivePrefix={arXiv},
      primaryClass={cs.CL},
      url={https://arxiv.org/abs/2605.22435}, 
}

@article{10.1093/pnasnexus/pgae111,
    author = {Mosleh, Mohsen and Cole, Rocky and Rand, David G},
    title = {Misinformation and harmful language are interconnected, rather than distinct, challenges},
    journal = {PNAS Nexus},
    volume = {3},
    number = {3},
    pages = {pgae111},
    year = {2024},
    month = {03},
    issn = {2752-6542},
    doi = {10.1093/pnasnexus/pgae111},
    url = {https://doi.org/10.1093/pnasnexus/pgae111},
    eprint = {https://academic.oup.com/pnasnexus/article-pdf/3/3/pgae111/57092205/pgae111.pdf},
}

@inproceedings{zheng-etal-2024-hatemoderate,
    title = "{H}ate{M}oderate: Testing Hate Speech Detectors against Content Moderation Policies",
    author = "Zheng, Jiangrui  and
      Liu, Xueqing  and
      Haque, Mirazul  and
      Qian, Xing  and
      Yang, Guanqun  and
      Yang, Wei",
    editor = "Duh, Kevin  and
      Gomez, Helena  and
      Bethard, Steven",
    booktitle = "Findings of the Association for Computational Linguistics: NAACL 2024",
    month = jun,
    year = "2024",
    address = "Mexico City, Mexico",
    publisher = "Association for Computational Linguistics",
    url = "https://aclanthology.org/2024.findings-naacl.172/",
    doi = "10.18653/v1/2024.findings-naacl.172",
    pages = "2691--2710"
}

@misc{bonaldi2026catchmeragdatasetcontextually,
      title={CATCH-ME if you RAG: a dataset of Contextually Annotated multi-Turn Counterspeech against Hate and Misinformation Exchanges}, 
      author={Helena Bonaldi and Genoveffa Martone and Marco Guerini},
      year={2026},
      eprint={2606.20369},
      archivePrefix={arXiv},
      primaryClass={cs.CL},
      url={https://arxiv.org/abs/2606.20369}, 
}

@book{doctorow2025enshittification,
  title={Enshittification: Why everything suddenly got worse and what to do about it},
  author={Doctorow, Cory},
  year={2025},
  publisher={Verso Books}
}

@misc{shaib2025measuringaisloptext,
      title={Measuring AI "Slop" in Text}, 
      author={Chantal Shaib and Tuhin Chakrabarty and Diego Garcia-Olano and Byron C. Wallace},
      year={2025},
      eprint={2509.19163},
      archivePrefix={arXiv},
      primaryClass={cs.CL},
      url={https://arxiv.org/abs/2509.19163}
}

@inproceedings{zhu2025judgelmfinetunedlargelanguage,
 author = {Zhu, Lianghui and Wang, Xinggang and Wang, Xinlong},
 booktitle = {International Conference on Learning Representations},
 editor = {Y. Yue and A. Garg and N. Peng and F. Sha and R. Yu},
 pages = {51257--51296},
 title = {JudgeLM: Fine-tuned Large Language Models are Scalable Judges},
 url = {https://proceedings.iclr.cc/paper_files/paper/2025/file/7f8f73134e253845a8f82983219a8452-Paper-Conference.pdf},
 volume = {2025},
 year = {2025}
}

@inproceedings{de-gibert-etal-2018-hate,
    title = "Hate Speech Dataset from a White Supremacy Forum",
    author = "de Gibert, Ona  and
      Perez, Naiara  and
      Garc{\'i}a-Pablos, Aitor  and
      Cuadros, Montse",
    editor = "Fi{\v{s}}er, Darja  and
      Huang, Ruihong  and
      Prabhakaran, Vinodkumar  and
      Voigt, Rob  and
      Waseem, Zeerak  and
      Wernimont, Jacqueline",
    booktitle = "Proceedings of the 2nd Workshop on Abusive Language Online ({ALW}2)",
    month = oct,
    year = "2018",
    address = "Brussels, Belgium",
    publisher = "Association for Computational Linguistics",
    url = "https://aclanthology.org/W18-5102/",
    doi = "10.18653/v1/W18-5102",
    pages = "11--20"
}

@techreport{vanDerLinden2023_APA_misinformation,
  title        = {Using Psychological Science to Understand and Fight Health Misinformation: An APA Consensus Statement},
  author       = {van der Linden, Sander and Albarrac{\'\i}n, Dolores and Fazio, Lisa and Freelon, Dominic and Roozenbeek, Johanna and Swire‐Thompson, Briony and Van Bavel, Jay},
  institution  = {APA},
  year         = {2023},
  url          = {https://www.apa.org/pubs/reports/misinformation-consensus-statement.pdf},
  note         = {APA consensus statement report}
}

@article{Wang2025,
author = {Xiaoou Wang and Elena Cabrio and Serena Villata},
title ={When automated fact-checking meets argumentation: Unveiling fake news through argumentative evidence},
journal = {Argument \& Computation},
volume = {16},
number = {3},
pages = {405-424},
year = {2025},
doi = {10.1177/19462174251330980},
URL = { 
        https://doi.org/10.1177/19462174251330980
},
eprint = { 
        https://doi.org/10.1177/19462174251330980
}
}

@inproceedings{russo-etal-2023-countering,
    title = "Countering Misinformation via Emotional Response Generation",
    author = "Russo, Daniel  and
      Kaszefski-Yaschuk, Shane  and
      Staiano, Jacopo  and
      Guerini, Marco",
    editor = "Bouamor, Houda  and
      Pino, Juan  and
      Bali, Kalika",
    booktitle = "Proceedings of the 2023 Conference on Empirical Methods in Natural Language Processing",
    month = dec,
    year = "2023",
    address = "Singapore",
    publisher = "Association for Computational Linguistics",
    url = "https://aclanthology.org/2023.emnlp-main.703/",
    doi = "10.18653/v1/2023.emnlp-main.703",
    pages = "11476--11492"
}

@article{russo-etal-2023-benchmarking,
    title = "Benchmarking the Generation of Fact Checking Explanations",
    author = "Russo, Daniel  and
      Tekiro{\u{g}}lu, Serra Sinem  and
      Guerini, Marco",
    journal = "Transactions of the Association for Computational Linguistics",
    volume = "11",
    year = "2023",
    address = "Cambridge, MA",
    publisher = "MIT Press",
    url = "https://aclanthology.org/2023.tacl-1.71/",
    doi = "10.1162/tacl_a_00601",
    pages = "1250--1264"
}

@misc{baumann2025largelanguagemodelhacking,
      title={Large Language Model Hacking: Quantifying the Hidden Risks of Using LLMs for Text Annotation}, 
      author={Joachim Baumann and Paul Röttger and Aleksandra Urman and Albert Wendsjö and Flor Miriam Plaza-del-Arco and Johannes B. Gruber and Dirk Hovy},
      year={2025},
      eprint={2509.08825},
      archivePrefix={arXiv},
      primaryClass={cs.CL},
      url={https://arxiv.org/abs/2509.08825}, 
}

@inproceedings{wang-etal-2023-primacy,
    title = "Primacy Effect of {C}hat{GPT}",
    author = "Wang, Yiwei  and
      Cai, Yujun  and
      Chen, Muhao  and
      Liang, Yuxuan  and
      Hooi, Bryan",
    editor = "Bouamor, Houda  and
      Pino, Juan  and
      Bali, Kalika",
    booktitle = "Proceedings of the 2023 Conference on Empirical Methods in Natural Language Processing",
    month = dec,
    year = "2023",
    address = "Singapore",
    publisher = "Association for Computational Linguistics",
    url = "https://aclanthology.org/2023.emnlp-main.8/",
    doi = "10.18653/v1/2023.emnlp-main.8",
    pages = "108--115"
}

@techreport{WardleDerakhshan2017,
  author       = {Claire Wardle and Hossein Derakhshan},
  title        = {Information Disorder: Toward an Interdisciplinary Framework for Research and Policy-making},
  institution  = {Council of Europe},
  type         = {Report (DGI(2017)091)},
  year         = {2017},
  month        = sep # "~27",
  address      = {Strasbourg, France},
  url          = {https://tverezo.info/wp-content/uploads/2017/11/PREMS-162317-GBR-2018-Report-desinformation-A4-BAT.pdf},
  note         = {© Council of Europe, October 2017}
}

@inproceedings{li-etal-2023-synthetic,
    title = "Synthetic Data Generation with Large Language Models for Text Classification: Potential and Limitations",
    author = "Li, Zhuoyan  and
      Zhu, Hangxiao  and
      Lu, Zhuoran  and
      Yin, Ming",
    editor = "Bouamor, Houda  and
      Pino, Juan  and
      Bali, Kalika",
    booktitle = "Proceedings of the 2023 Conference on Empirical Methods in Natural Language Processing",
    month = dec,
    year = "2023",
    address = "Singapore",
    publisher = "Association for Computational Linguistics",
    url = "https://aclanthology.org/2023.emnlp-main.647/",
    doi = "10.18653/v1/2023.emnlp-main.647",
    pages = "10443--10461"
}

@inproceedings{gpt-4-as-annotator,
author = {He, Zeyu and Huang, Chieh-Yang and Ding, Chien-Kuang Cornelia and Rohatgi, Shaurya and Huang, Ting-Hao Kenneth},
title = {If in a Crowdsourced Data Annotation Pipeline, a GPT-4},
year = {2024},
isbn = {9798400703300},
publisher = {Association for Computing Machinery},
address = {New York, NY, USA},
url = {https://doi.org/10.1145/3613904.3642834},
doi = {10.1145/3613904.3642834},
booktitle = {Proceedings of the 2024 CHI Conference on Human Factors in Computing Systems},
articleno = {1040},
numpages = {25},
location = {Honolulu, HI, USA},
series = {CHI '24}
}

@article{ziems-etal-2024-large,
    title = "Can Large Language Models Transform Computational Social Science?",
    author = "Ziems, Caleb  and
      Held, William  and
      Shaikh, Omar  and
      Chen, Jiaao  and
      Zhang, Zhehao  and
      Yang, Diyi",
    journal = "Computational Linguistics",
    volume = "50",
    number = "1",
    month = mar,
    year = "2024",
    address = "Cambridge, MA",
    publisher = "MIT Press",
    url = "https://aclanthology.org/2024.cl-1.8/",
    doi = "10.1162/coli_a_00502",
    pages = "237--291"
}

@inproceedings{wang-etal-2021-want-reduce,
    title = "Want To Reduce Labeling Cost? {GPT}-3 Can Help",
    author = "Wang, Shuohang  and
      Liu, Yang  and
      Xu, Yichong  and
      Zhu, Chenguang  and
      Zeng, Michael",
    editor = "Moens, Marie-Francine  and
      Huang, Xuanjing  and
      Specia, Lucia  and
      Yih, Scott Wen-tau",
    booktitle = "Findings of the Association for Computational Linguistics: EMNLP 2021",
    month = nov,
    year = "2021",
    address = "Punta Cana, Dominican Republic",
    publisher = "Association for Computational Linguistics",
    url = "https://aclanthology.org/2021.findings-emnlp.354/",
    doi = "10.18653/v1/2021.findings-emnlp.354",
    pages = "4195--4205"
}

@inproceedings{lin-etal-2023-argue,
    title = "Argue with Me Tersely: Towards Sentence-Level Counter-Argument Generation",
    author = "Lin, Jiayu  and
      Ye, Rong  and
      Han, Meng  and
      Zhang, Qi  and
      Lai, Ruofei  and
      Zhang, Xinyu  and
      Cao, Zhao  and
      Huang, Xuanjing  and
      Wei, Zhongyu",
    editor = "Bouamor, Houda  and
      Pino, Juan  and
      Bali, Kalika",
    booktitle = "Proceedings of the 2023 Conference on Empirical Methods in Natural Language Processing",
    month = dec,
    year = "2023",
    address = "Singapore",
    publisher = "Association for Computational Linguistics",
    url = "https://aclanthology.org/2023.emnlp-main.1039/",
    doi = "10.18653/v1/2023.emnlp-main.1039",
    pages = "16705--16720"
}

@inproceedings{bonaldi-etal-2025-first,
    title = "The First Workshop on Multilingual Counterspeech Generation at {COLING} 2025: Overview of the Shared Task",
    author = "Bonaldi, Helena  and
      Vallecillo-Rodr{\'i}guez, Mar{\'i}a Estrella  and
      Zubiaga, Irune  and
      Montejo-Raez, Arturo  and
      Soroa, Aitor  and
      Mart{\'i}n-Valdivia, Mar{\'i}a-Teresa  and
      Guerini, Marco  and
      Agerri, Rodrigo",
    editor = "Bonaldi, Helena  and
      Vallecillo-Rodr{\'i}guez, Mar{\'i}a Estrella  and
      Zubiaga, Irune  and
      Montejo-R{\'a}ez, Arturo  and
      Soroa, Aitor  and
      Mart{\'i}n-Valdivia, Mar{\'i}a Teresa  and
      Guerini, Marco  and
      Agerri, Rodrigo",
    booktitle = "Proceedings of the First Workshop on Multilingual Counterspeech Generation",
    month = jan,
    year = "2025",
    address = "Abu Dhabi, UAE",
    publisher = "Association for Computational Linguistics",
    url = "https://aclanthology.org/2025.mcg-1.10/",
    pages = "92--107"
}

@inproceedings{zubiaga-etal-2024-llm,
    title = "A {LLM}-based Ranking Method for the Evaluation of Automatic Counter-Narrative Generation",
    author = "Zubiaga, Irune  and
      Soroa, Aitor  and
      Agerri, Rodrigo",
    editor = "Al-Onaizan, Yaser  and
      Bansal, Mohit  and
      Chen, Yun-Nung",
    booktitle = "Findings of the Association for Computational Linguistics: EMNLP 2024",
    month = nov,
    year = "2024",
    address = "Miami, Florida, USA",
    publisher = "Association for Computational Linguistics",
    url = "https://aclanthology.org/2024.findings-emnlp.559/",
    doi = "10.18653/v1/2024.findings-emnlp.559",
    pages = "9572--9585"
}

@inproceedings{claimbuster,
author = {Hassan, Naeemul and Li, Chengkai and Tremayne, Mark},
title = {Detecting Check-worthy Factual Claims in Presidential Debates},
year = {2015},
isbn = {9781450337946},
publisher = {Association for Computing Machinery},
address = {New York, NY, USA},
url = {https://doi.org/10.1145/2806416.2806652},
doi = {10.1145/2806416.2806652},
booktitle = {Proceedings of the 24th ACM International on Conference on Information and Knowledge Management},
pages = {1835–1838},
numpages = {4},
location = {Melbourne, Australia},
series = {CIKM '15}
}

@inproceedings{murzaku2022re,
    title = "Re-Examining {F}act{B}ank: Predicting the Author{'}s Presentation of Factuality",
    author = "Murzaku, John  and
      Zeng, Peter  and
      Markowska, Magdalena  and
      Rambow, Owen",
    editor = "Calzolari, Nicoletta  and
      Huang, Chu-Ren  and
      Kim, Hansaem  and
      Pustejovsky, James  and
      Wanner, Leo  and
      Choi, Key-Sun  and
      Ryu, Pum-Mo  and
      Chen, Hsin-Hsi  and
      Donatelli, Lucia  and
      Ji, Heng  and
      Kurohashi, Sadao  and
      Paggio, Patrizia  and
      Xue, Nianwen  and
      Kim, Seokhwan  and
      Hahm, Younggyun  and
      He, Zhong  and
      Lee, Tony Kyungil  and
      Santus, Enrico  and
      Bond, Francis  and
      Na, Seung-Hoon",
    booktitle = "Proceedings of the 29th International Conference on Computational Linguistics",
    month = oct,
    year = "2022",
    address = "Gyeongju, Republic of Korea",
    publisher = "International Committee on Computational Linguistics",
    url = "https://aclanthology.org/2022.coling-1.66/",
    pages = "786--796"
}

@inproceedings{fanton-etal-2021-human,
    title = "Human-in-the-Loop for Data Collection: a Multi-Target Counter Narrative Dataset to Fight Online Hate Speech",
    author = "Fanton, Margherita  and
      Bonaldi, Helena  and
      Tekiro{\u{g}}lu, Serra Sinem  and
      Guerini, Marco",
    editor = "Zong, Chengqing  and
      Xia, Fei  and
      Li, Wenjie  and
      Navigli, Roberto",
    booktitle = "Proceedings of the 59th Annual Meeting of the Association for Computational Linguistics and the 11th International Joint Conference on Natural Language Processing (Volume 1: Long Papers)",
    month = aug,
    year = "2021",
    address = "Online",
    publisher = "Association for Computational Linguistics",
    url = "https://aclanthology.org/2021.acl-long.250/",
    doi = "10.18653/v1/2021.acl-long.250",
    pages = "3226--3240"
}

@inproceedings{chung-etal-2019-conan,
    title = "{CONAN} - {CO}unter {NA}rratives through Nichesourcing: a Multilingual Dataset of Responses to Fight Online Hate Speech",
    author = "Chung, Yi-Ling  and
      Kuzmenko, Elizaveta  and
      Tekiroglu, Serra Sinem  and
      Guerini, Marco",
    editor = "Korhonen, Anna  and
      Traum, David  and
      M{\`a}rquez, Llu{\'i}s",
    booktitle = "Proceedings of the 57th Annual Meeting of the Association for Computational Linguistics",
    month = jul,
    year = "2019",
    address = "Florence, Italy",
    publisher = "Association for Computational Linguistics",
    url = "https://aclanthology.org/P19-1271/",
    doi = "10.18653/v1/P19-1271",
    pages = "2819--2829"
}

@inproceedings{bonaldi-etal-2024-safer,
    title = "Is Safer Better? The Impact of Guardrails on the Argumentative Strength of {LLM}s in Hate Speech Countering",
    author = "Bonaldi, Helena  and
      Damo, Greta  and
      Ocampo, Nicol{\'a}s Benjam{\'i}n  and
      Cabrio, Elena  and
      Villata, Serena  and
      Guerini, Marco",
    editor = "Al-Onaizan, Yaser  and
      Bansal, Mohit  and
      Chen, Yun-Nung",
    booktitle = "Proceedings of the 2024 Conference on Empirical Methods in Natural Language Processing",
    month = nov,
    year = "2024",
    address = "Miami, Florida, USA",
    publisher = "Association for Computational Linguistics",
    url = "https://aclanthology.org/2024.emnlp-main.201/",
    doi = "10.18653/v1/2024.emnlp-main.201",
    pages = "3446--3463"
}

@inproceedings{ocampo-etal-2023-depth,
    title = "An In-depth Analysis of Implicit and Subtle Hate Speech Messages",
    author = "Ocampo, Nicol{\'a}s Benjam{\'i}n  and
      Sviridova, Ekaterina  and
      Cabrio, Elena  and
      Villata, Serena",
    editor = "Vlachos, Andreas  and
      Augenstein, Isabelle",
    booktitle = "Proceedings of the 17th Conference of the European Chapter of the Association for Computational Linguistics",
    month = may,
    year = "2023",
    address = "Dubrovnik, Croatia",
    publisher = "Association for Computational Linguistics",
    url = "https://aclanthology.org/2023.eacl-main.147/",
    doi = "10.18653/v1/2023.eacl-main.147",
    pages = "1997--2013"
}

@inproceedings{hartvigsen-etal-2022-toxigen,
    title = "{T}oxi{G}en: A Large-Scale Machine-Generated Dataset for Adversarial and Implicit Hate Speech Detection",
    author = "Hartvigsen, Thomas  and
      Gabriel, Saadia  and
      Palangi, Hamid  and
      Sap, Maarten  and
      Ray, Dipankar  and
      Kamar, Ece",
    editor = "Muresan, Smaranda  and
      Nakov, Preslav  and
      Villavicencio, Aline",
    booktitle = "Proceedings of the 60th Annual Meeting of the Association for Computational Linguistics (Volume 1: Long Papers)",
    month = may,
    year = "2022",
    address = "Dublin, Ireland",
    publisher = "Association for Computational Linguistics",
    url = "https://aclanthology.org/2022.acl-long.234/",
    doi = "10.18653/v1/2022.acl-long.234",
    pages = "3309--3326"
}

@inproceedings{vidgen-etal-2021-learning,
    title = "Learning from the Worst: Dynamically Generated Datasets to Improve Online Hate Detection",
    author = "Vidgen, Bertie  and
      Thrush, Tristan  and
      Waseem, Zeerak  and
      Kiela, Douwe",
    editor = "Zong, Chengqing  and
      Xia, Fei  and
      Li, Wenjie  and
      Navigli, Roberto",
    booktitle = "Proceedings of the 59th Annual Meeting of the Association for Computational Linguistics and the 11th International Joint Conference on Natural Language Processing (Volume 1: Long Papers)",
    month = aug,
    year = "2021",
    address = "Online",
    publisher = "Association for Computational Linguistics",
    url = "https://aclanthology.org/2021.acl-long.132/",
    doi = "10.18653/v1/2021.acl-long.132",
    pages = "1667--1682"
}

@inproceedings{sap-etal-2020-social,
    title = "Social Bias Frames: Reasoning about Social and Power Implications of Language",
    author = "Sap, Maarten  and
      Gabriel, Saadia  and
      Qin, Lianhui  and
      Jurafsky, Dan  and
      Smith, Noah A.  and
      Choi, Yejin",
    editor = "Jurafsky, Dan  and
      Chai, Joyce  and
      Schluter, Natalie  and
      Tetreault, Joel",
    booktitle = "Proceedings of the 58th Annual Meeting of the Association for Computational Linguistics",
    month = jul,
    year = "2020",
    address = "Online",
    publisher = "Association for Computational Linguistics",
    url = "https://aclanthology.org/2020.acl-main.486/",
    doi = "10.18653/v1/2020.acl-main.486",
    pages = "5477--5490"
}

@inproceedings{elsherief-etal-2021-latent,
    title = "Latent Hatred: A Benchmark for Understanding Implicit Hate Speech",
    author = "ElSherief, Mai  and
      Ziems, Caleb  and
      Muchlinski, David  and
      Anupindi, Vaishnavi  and
      Seybolt, Jordyn  and
      De Choudhury, Munmun  and
      Yang, Diyi",
    editor = "Moens, Marie-Francine  and
      Huang, Xuanjing  and
      Specia, Lucia  and
      Yih, Scott Wen-tau",
    booktitle = "Proceedings of the 2021 Conference on Empirical Methods in Natural Language Processing",
    month = nov,
    year = "2021",
    address = "Online and Punta Cana, Dominican Republic",
    publisher = "Association for Computational Linguistics",
    url = "https://aclanthology.org/2021.emnlp-main.29/",
    doi = "10.18653/v1/2021.emnlp-main.29",
    pages = "345--363"
}

@inproceedings{Davidson2017,
  author       = {Thomas Davidson and
                  Dana Warmsley and
                  Michael W. Macy and
                  Ingmar Weber},
  title        = {Automated Hate Speech Detection and the Problem of Offensive Language},
  booktitle    = {Proceedings of the Eleventh International Conference on Web and Social
                  Media, {ICWSM} 2017, Montr{\'{e}}al, Qu{\'{e}}bec, Canada,
                  May 15-18, 2017},
  pages        = {512--515},
  publisher    = {{AAAI} Press},
  year         = {2017},
  url          = {https://aaai.org/ocs/index.php/ICWSM/ICWSM17/paper/view/15665},
  bibsource    = {dblp computer science bibliography, https://dblp.org}
}

@inproceedings{zampieri-etal-2020-semeval,
    title = "{S}em{E}val-2020 Task 12: Multilingual Offensive Language Identification in Social Media ({O}ffens{E}val 2020)",
    author = {Zampieri, Marcos  and
      Nakov, Preslav  and
      Rosenthal, Sara  and
      Atanasova, Pepa  and
      Karadzhov, Georgi  and
      Mubarak, Hamdy  and
      Derczynski, Leon  and
      Pitenis, Zeses  and
      {\c{C}}{\"o}ltekin, {\c{C}}a{\u{g}}r{\i}},
    editor = "Herbelot, Aurelie  and
      Zhu, Xiaodan  and
      Palmer, Alexis  and
      Schneider, Nathan  and
      May, Jonathan  and
      Shutova, Ekaterina",
    booktitle = "Proceedings of the Fourteenth Workshop on Semantic Evaluation",
    month = dec,
    year = "2020",
    address = "Barcelona (online)",
    publisher = "International Committee for Computational Linguistics",
    url = "https://aclanthology.org/2020.semeval-1.188/",
    doi = "10.18653/v1/2020.semeval-1.188",
    pages = "1425--1447"
}

@inproceedings{basile-etal-2019-semeval,
    title = "{S}em{E}val-2019 Task 5: Multilingual Detection of Hate Speech Against Immigrants and Women in {T}witter",
    author = "Basile, Valerio  and
      Bosco, Cristina  and
      Fersini, Elisabetta  and
      Nozza, Debora  and
      Patti, Viviana  and
      Rangel Pardo, Francisco Manuel  and
      Rosso, Paolo  and
      Sanguinetti, Manuela",
    editor = "May, Jonathan  and
      Shutova, Ekaterina  and
      Herbelot, Aurelie  and
      Zhu, Xiaodan  and
      Apidianaki, Marianna  and
      Mohammad, Saif M.",
    booktitle = "Proceedings of the 13th International Workshop on Semantic Evaluation",
    month = jun,
    year = "2019",
    address = "Minneapolis, Minnesota, USA",
    publisher = "Association for Computational Linguistics",
    url = "https://aclanthology.org/S19-2007/",
    doi = "10.18653/v1/S19-2007",
    pages = "54--63"
}

@inproceedings{jaradat-etal-2018-claimrank,
    title = "{C}laim{R}ank: Detecting Check-Worthy Claims in {A}rabic and {E}nglish",
    author = "Jaradat, Israa  and
      Gencheva, Pepa  and
      Barr{\'o}n-Cede{\~n}o, Alberto  and
      M{\`a}rquez, Llu{\'i}s  and
      Nakov, Preslav",
    editor = "Liu, Yang  and
      Paek, Tim  and
      Patwardhan, Manasi",
    booktitle = "Proceedings of the 2018 Conference of the North {A}merican Chapter of the Association for Computational Linguistics: Demonstrations",
    month = jun,
    year = "2018",
    address = "New Orleans, Louisiana",
    publisher = "Association for Computational Linguistics",
    url = "https://aclanthology.org/N18-5006/",
    doi = "10.18653/v1/N18-5006",
    pages = "26--30"
}

@inproceedings{wright-augenstein-2020-claim,
    title = "Claim Check-Worthiness Detection as Positive Unlabelled Learning",
    author = "Wright, Dustin  and
      Augenstein, Isabelle",
    editor = "Cohn, Trevor  and
      He, Yulan  and
      Liu, Yang",
    booktitle = "Findings of the Association for Computational Linguistics: EMNLP 2020",
    month = nov,
    year = "2020",
    address = "Online",
    publisher = "Association for Computational Linguistics",
    url = "https://aclanthology.org/2020.findings-emnlp.43/",
    doi = "10.18653/v1/2020.findings-emnlp.43",
    pages = "476--488"
}

@inproceedings{ijcai2021p619,
  title     = {Automated Fact-Checking for Assisting Human Fact-Checkers},
  author    = {Nakov, Preslav and Corney, David and Hasanain, Maram and Alam, Firoj and Elsayed, Tamer and Barrón-Cedeño, Alberto and Papotti, Paolo and Shaar, Shaden and Da San Martino, Giovanni},
  booktitle = {Proceedings of the Thirtieth International Joint Conference on
               Artificial Intelligence, {IJCAI-21}},
  publisher = {International Joint Conferences on Artificial Intelligence Organization},
  editor    = {Zhi-Hua Zhou},
  pages     = {4551--4558},
  year      = {2021},
  month     = {8},
  note      = {Survey Track},
  doi       = {10.24963/ijcai.2021/619},
  url       = {https://doi.org/10.24963/ijcai.2021/619},
}

@inproceedings{renze-2024-effect,
    title = "The Effect of Sampling Temperature on Problem Solving in Large Language Models",
    author = "Renze, Matthew",
    editor = "Al-Onaizan, Yaser  and
      Bansal, Mohit  and
      Chen, Yun-Nung",
    booktitle = "Findings of the Association for Computational Linguistics: EMNLP 2024",
    month = nov,
    year = "2024",
    address = "Miami, Florida, USA",
    publisher = "Association for Computational Linguistics",
    url = "https://aclanthology.org/2024.findings-emnlp.432/",
    doi = "10.18653/v1/2024.findings-emnlp.432",
    pages = "7346--7356"
}

@article{LI2025242,
title = {Exploring the Impact of Temperature on Large Language Models: Hot or Cold?},
journal = {Procedia Computer Science},
volume = {264},
pages = {242-251},
year = {2025},
note = {International Neural Network Society Workshop on Deep Learning Innovations and Applications 2025},
issn = {1877-0509},
doi = {https://doi.org/10.1016/j.procs.2025.07.135},
url = {https://www.sciencedirect.com/science/article/pii/S1877050925021854},
author = {Lujun Li and Lama Sleem and Niccolo’ Gentile and Geoffrey Nichil and Radu State}
}

@inproceedings{alam-etal-2021-fighting-covid,
    title = "Fighting the {COVID}-19 Infodemic: Modeling the Perspective of Journalists, Fact-Checkers, Social Media Platforms, Policy Makers, and the Society",
    author = "Alam, Firoj  and
      Shaar, Shaden  and
      Dalvi, Fahim  and
      Sajjad, Hassan  and
      Nikolov, Alex  and
      Mubarak, Hamdy  and
      Da San Martino, Giovanni  and
      Abdelali, Ahmed  and
      Durrani, Nadir  and
      Darwish, Kareem  and
      Al-Homaid, Abdulaziz  and
      Zaghouani, Wajdi  and
      Caselli, Tommaso  and
      Danoe, Gijs  and
      Stolk, Friso  and
      Bruntink, Britt  and
      Nakov, Preslav",
    editor = "Moens, Marie-Francine  and
      Huang, Xuanjing  and
      Specia, Lucia  and
      Yih, Scott Wen-tau",
    booktitle = "Findings of the Association for Computational Linguistics: EMNLP 2021",
    month = nov,
    year = "2021",
    address = "Punta Cana, Dominican Republic",
    publisher = "Association for Computational Linguistics",
    url = "https://aclanthology.org/2021.findings-emnlp.56/",
    doi = "10.18653/v1/2021.findings-emnlp.56",
    pages = "611--649"
}

@inproceedings{majer-snajder-2024-claim,
    title = "Claim Check-Worthiness Detection: How Well do {LLM}s Grasp Annotation Guidelines?",
    author = "Majer, Laura  and
      {\v{S}}najder, Jan",
    editor = "Schlichtkrull, Michael  and
      Chen, Yulong  and
      Whitehouse, Chenxi  and
      Deng, Zhenyun  and
      Akhtar, Mubashara  and
      Aly, Rami  and
      Guo, Zhijiang  and
      Christodoulopoulos, Christos  and
      Cocarascu, Oana  and
      Mittal, Arpit  and
      Thorne, James  and
      Vlachos, Andreas",
    booktitle = "Proceedings of the Seventh Fact Extraction and VERification Workshop (FEVER)",
    month = nov,
    year = "2024",
    address = "Miami, Florida, USA",
    publisher = "Association for Computational Linguistics",
    url = "https://aclanthology.org/2024.fever-1.27/",
    doi = "10.18653/v1/2024.fever-1.27",
    pages = "245--263"
}

@inproceedings{shaar2021overview,
  author       = {Shaden Shaar and
                  Maram Hasanain and
                  Bayan Hamdan and
                  Zien Sheikh Ali and
                  Fatima Haouari and
                  Alex Nikolov and
                  M{\"{u}}cahid Kutlu and
                  Yavuz Selim Kartal and
                  Firoj Alam and
                  Giovanni Da San Martino and
                  Alberto Barr{\'{o}}n{-}Cede{\~{n}}o and
                  Rub{\'{e}}n M{\'{\i}}guez and
                  Javier Beltr{\'{a}}n and
                  Tamer Elsayed and
                  Preslav Nakov},
  editor       = {Guglielmo Faggioli and
                  Nicola Ferro and
                  Alexis Joly and
                  Maria Maistro and
                  Florina Piroi},
  title        = {Overview of the {CLEF-2021} CheckThat! Lab Task 1 on Check-Worthiness
                  Estimation in Tweets and Political Debates},
  booktitle    = {Proceedings of the Working Notes of {CLEF} 2021 - Conference and Labs
                  of the Evaluation Forum, Bucharest, Romania, September 21st - to -
                  24th, 2021},
  series       = {{CEUR} Workshop Proceedings},
  volume       = {2936},
  pages        = {369--392},
  publisher    = {CEUR-WS.org},
  year         = {2021},
  url          = {https://ceur-ws.org/Vol-2936/paper-28.pdf},
  bibsource    = {dblp computer science bibliography, https://dblp.org}
}

@article{guo-etal-2022-survey,
    title = "A Survey on Automated Fact-Checking",
    author = "Guo, Zhijiang  and
      Schlichtkrull, Michael  and
      Vlachos, Andreas",
    editor = "Roark, Brian  and
      Nenkova, Ani",
    journal = "Transactions of the Association for Computational Linguistics",
    volume = "10",
    year = "2022",
    address = "Cambridge, MA",
    publisher = "MIT Press",
    url = "https://aclanthology.org/2022.tacl-1.11/",
    doi = "10.1162/tacl_a_00454",
    pages = "178--206"
}

@inproceedings{zerong-etal-2025-systematic,
    title = "A Systematic Survey of Claim Verification: Corpora, Systems, and Case Studies",
    author = "Zerong, Zhaxi  and
      Li, Chenxi  and
      Liu, Xinyi  and
      Chen, Ju-hui  and
      Xia, Fei",
    editor = "Christodoulopoulos, Christos  and
      Chakraborty, Tanmoy  and
      Rose, Carolyn  and
      Peng, Violet",
    booktitle = "Findings of the Association for Computational Linguistics: EMNLP 2025",
    month = nov,
    year = "2025",
    address = "Suzhou, China",
    publisher = "Association for Computational Linguistics",
    url = "https://aclanthology.org/2025.findings-emnlp.1170/",
    doi = "10.18653/v1/2025.findings-emnlp.1170",
    pages = "21452--21474",
    ISBN = "979-8-89176-335-7"
}

@inproceedings{hardalov-etal-2022-survey,
    title = "A Survey on Stance Detection for Mis- and Disinformation Identification",
    author = "Hardalov, Momchil  and
      Arora, Arnav  and
      Nakov, Preslav  and
      Augenstein, Isabelle",
    editor = "Carpuat, Marine  and
      de Marneffe, Marie-Catherine  and
      Meza Ruiz, Ivan Vladimir",
    booktitle = "Findings of the Association for Computational Linguistics: NAACL 2022",
    month = jul,
    year = "2022",
    address = "Seattle, United States",
    publisher = "Association for Computational Linguistics",
    url = "https://aclanthology.org/2022.findings-naacl.94/",
    doi = "10.18653/v1/2022.findings-naacl.94",
    pages = "1259--1277"
}

@inproceedings{rottger-etal-2022-two,
    title = "Two Contrasting Data Annotation Paradigms for Subjective {NLP} Tasks",
    author = {R{\"o}ttger, Paul  and
      Vidgen, Bertie  and
      Hovy, Dirk  and
      Pierrehumbert, Janet},
    editor = "Carpuat, Marine  and
      de Marneffe, Marie-Catherine  and
      Meza Ruiz, Ivan Vladimir",
    booktitle = "Proceedings of the 2022 Conference of the North American Chapter of the Association for Computational Linguistics: Human Language Technologies",
    month = jul,
    year = "2022",
    address = "Seattle, United States",
    publisher = "Association for Computational Linguistics",
    url = "https://aclanthology.org/2022.naacl-main.13/",
    doi = "10.18653/v1/2022.naacl-main.13",
    pages = "175--190"
}

\vfill \break 
\clearpage
\newpage
\appendix

\setcounter{table}{0}
\renewcommand{\tablename}{Table} 
\renewcommand{\thetable}{\Alph{table}}

\section{Specific Version of LLMs Evaluated}
\label{sec:appendix-llms-version}
We use the following open-access Instruct-based models from Hugging Face \footnote{\url{https://huggingface.co/models}}. All models are publicly available and traceable through their repository identifiers:

\noindent \textbf{Small Size Models}:
\begin{itemize}
    \item Mistral-7B: \texttt{mistralai/} \\ \texttt{Mistral-7B-Instruct-v0.3}
    \item Llama-8B: \texttt{meta-llama/} \\ \texttt{Llama-3.1-8B-Instruct}
    \item Olmo2-7B: \texttt{allenai/} \\ \texttt{OLMo-2-1124-7B-Instruct}
    \item Qwen2.5-7B: \texttt{Qwen/Qwen2.5-7B-Instruct}
    \item Command-r-7B: \texttt{CohereLabs/} \\ \texttt{c4ai-command-r7b-12-2024}
\end{itemize}

\noindent \textbf{Medium Size Models}:
\begin{itemize}
    \item Mixtral-8x7B: \texttt{mistralai/} \\ \texttt{Mixtral-8x7B-Instruct-v0.1}
    \item Mistral-22B: \texttt{mistralai/} \\ \texttt{Mistral-Small-Instruct-2409}
    \item Olmo2-32B: \texttt{allenai/} \\ \texttt{OLMo-2-0325-32B-Instruct}
    \item Mixtral-8x22B: \texttt{mistralai/} \\ \texttt{Mixtral-8x22B-Instruct-v0.1}
\end{itemize}

\noindent \textbf{Large Size Models}:
\begin{itemize}
    \item Llama-70B: \texttt{meta-llama/} \\ \texttt{Llama-3.3-70B-Instruct}
    \item Qwen2.5-72B: \texttt{Qwen/} \\ \texttt{Qwen2.5-72B-Instruct}
    \item Command-R-104B: \texttt{CohereLabs/} \\ \texttt{c4ai-command-r-plus-08-2024}
\end{itemize}

\section{Additional Details on the LLM-in-the-loop Evaluation}
\label{sec:add-details-llm-loop}

In this section, we describe the prompts used for check-worthiness detection (Section~\ref{sec:prompts-cw}). Next, we report the inter-annotator agreement across annotator pairs for the full human annotation of WSF-ARG+ (Section~\ref{sec:iaa-all}). Finally, we present the prediction variability of LLMs across three runs per configuration in the LLM-in-the-loop framework for check-worthiness detection (Section~\ref{sec:pred-var}).

\subsection{Prompts for the Check-worthiness Detection Task}
\label{sec:prompts-cw}

We tested two prompt strategies, zero-shot and one-shot, using a system–user prompt setup with structured outputs through the vLLM Python library \footnote{\url{https://docs.vllm.ai/en/latest/}}:

\noindent \textbf{System Prompt}: \texttt{You are an expert annotator that classifies text based on their check-worthiness. Always follow the definitions exactly.}

\noindent \textbf{User Prompt}: \texttt{Classify the following text into one of these categories:  \\
1.\textbf{Non-Factual}: Subjective text such as opinions, beliefs, declarations, or wishes. Many questions also fall into this category. These sentences do not contain any factual claim. \\
2.\textbf{Unimportant Factual}: Text that contain factual claims but are not important for fact-checking. The general public would not be interested in verifying them.  \\
3.\textbf{Check-worthy Factual}: Text that contain factual claims of public interest. These are the kinds of claims journalists would fact-check.  \\
Input text: `\{input\_text\}'
}

\noindent \textbf{Output Labels}: \texttt{["Non-Factual", "Unimportant Factual", "Check-worthy Factual"]}

Where \texttt{input\_text} was the claim from WSF-ARG+ to be annotated.

For the one-shot experiments we used one example per check-worthiness label appended after the definition of each class. We used the same examples for all the predicted claims in WSF-ARG+. Examples are extracted from the dataset.

\paragraph{CFS One Shot Example:} \textit{“92\% of abortion clinics are in black communities.”}

\paragraph{UFS One Shot Example:} \textit{"When I was last in South Africa, we went to Sun City and saw little black monkey children swimming.”}

\paragraph{NFS One Shot Example:} \textit{"We should go to the southwest and go on the offensive and drive every race down into Mexico.”}

\subsection{Inter-Annotator Agreement of the Human Evaluation Between Annotator Pairs}
\label{sec:iaa-all}

As described in Section \ref{sec:human-val}, the entire dataset was annotated by three annotators. To measure inter-annotator agreement, we first compute pairwise agreement using Cohen’s $\kappa$ for the check-worthiness detection task, considering both the three-label scheme—NFS, UFS, CFS (CW-3)—and the two-label scheme, where NFS and UFS are merged as Non-Check-worthy and CFS as Check-worthy (CW-2). For agreement across all three annotators, we calculate Krippendorff’s $\alpha$, which supports more than two raters. Table~\ref{tab:human-eval-iaa} shows that Cohen's $\kappa$ agreement between annotator pairs ranges from 0.485 to 0.627 for both CW-3 and CW-2, while Krippendorff's $\alpha$ agreement among the three raters is 0.537 for CW-3 and 0.570 for CW-2. Overall, both pairwise and three-way agreement indicate moderate to substantial agreement, even for a highly subjective task such as detecting check-worthiness in the context of hate speech. 

\begin{table}[t]
\begin{tabular}{cp{2cm}p{2cm}}
\toprule
 & \multicolumn{2}{l}{Inter-Annotator Agreement}
 \\ \midrule
Pair                              & CW-3 classes & CW-2 classes \\ \midrule
Ann1 vs Ann2                      & 0.571                    & 0.627                                       \\
Ann1 vs Ann3                      & 0.506                    & 0.545                                       \\
Ann2 vs Ann3                      & 0.485                    & 0.547                                       \\ \midrule
All 3 Annotators                  & 0.537                    & 0.570                             \\ \bottomrule         
\end{tabular}
\caption{IAA for WSF-ARG+, full human annotation: CW-3 (NFS, UFS, CFS) and CW-2 (CFS vs. Non-Check-worthy, merging NFS and UFS). Pairwise IAA: Cohen’s $\kappa$; all three annotators: Krippendorff’s $\alpha$. Ann1-Ann3: Annotators one to three of the task.}
\label{tab:human-eval-iaa}
\end{table}

\subsection{Prediction Variability}
\label{sec:pred-var}

For each LLM and prompt (zero-shot or one-shot), we ran three independent predictions per claim in  WSF-ARG+. Table \ref{tab:variability} shows the consistency of these predictions by reporting the proportion of cases where all three runs produced the same label (“all equal”), where two runs agreed and one disagreed (“2 equal”), and where all three runs produced different labels (“unequal”). The results show that fewer than 3\% of cases required human intervention due to unstable labeling, and that one-shot prompting increases stability by reducing the proportion of “2 equal” and “unequal” cases.

\begin{table}[t]
\centering
\resizebox{!}{3.4cm}{\
\begin{tabular}{
l 
l|
c 
c 
c 
c 
c 
c|
c
c 
c 
c 
c 
c}
\toprule
                                  &           & \multicolumn{6}{c|}{HS Messages} & \multicolumn{6}{c}{Non-HS Messages}                                                                  \\ \cmidrule{3-14}
                                                    &     & \multicolumn{2}{c}{\text{all equal}} & \multicolumn{2}{c}{\text{2 equal}}  & \multicolumn{2}{c|}{\text{unequal} } & \multicolumn{2}{c}{\text{all equal}} & \multicolumn{2}{c}{\text{2 equal}} & \multicolumn{2}{c}{\text{unequal} } \\ \cmidrule{3-14}
Model                                 & Shot     & \# & \% & \# &\% & \# & \% & \# & \% & \# &\% & \# & \%  \\ \midrule
& zero & 632 & 99.8 & 1   & 0.2	& 0.0 & 0.0 & 965 & 98.3 & 17  & 1.7    & 0	 & 0.0 \\
\multirow{-2}{*}{Mistral-7B}     & one  & 612 & 96.7 & 21  & 3.3	& 0.0 & 0.0 & 977 & 99.5 & 5   & 0.5    & 0	 & 0.0 \\
& zero & 614 & 97.0 & 19  & 3.0	& 0.0 & 0.0 & 897 & 91.3 & 84  & 8.6    & 1	 & 0.1 \\
\multirow{-2}{*}{Llama-8B}       & one  & 632 & 99.8 & 1   & 0.2	& 0.0 & 0.0 & 955 & 97.3 & 27  & 2.7    & 0	 & 0.0 \\
& zero & 520 & 82.1 & 113 & 17.9	& 0.0 & 0.0 & 896 & 91.2 & 82  & 8.4    & 4	 & 0.4 \\
\multirow{-2}{*}{Olmo2-7B}       & one  & 572 & 90.4 & 61  & 9.6	& 0.0 & 0.0 & 861 & 87.7 & 121 & 12.3   & 0	 & 0.0 \\
& zero & 618 & 97.6 & 15  & 2.4	& 0.0 & 0.0 & 938 & 95.5 & 44  & 4.5    & 0	 & 0.0 \\
\multirow{-2}{*}{Qwen2.5-7B}     & one  & 619 & 97.8 & 14  & 2.2	& 0.0 & 0.0 & 976 & 99.4 & 6   & 0.6    & 0	 & 0.0 \\
& zero & 492 & 77.7 & 140 & 22.1	& 1.0 & 0.2 & 694 & 70.7 & 280 & 28.5   & 8	 & 0.8 \\
\multirow{-2}{*}{Com.-r-7B}      & one  & 510 & 80.6 & 123 & 19.4	& 0.0 & 0.0 & 761 & 77.5 & 219 & 22.3   & 2	 & 0.2 \\ \midrule
& zero & 517 & 81.7 & 112 & 17.7	& 4.0 & 0.6 & 699 & 71.2 & 254 & 25.9   & 29 & 3.0 \\
\multirow{-2}{*}{Mixtral-8x7B}   & one  & 531 & 83.9 & 93  & 14.7	& 9.0 & 1.4 & 774 & 78.8 & 185 & 18.8   & 23 & 2.3 \\
& zero & 610 & 96.4 & 23  & 3.6	& 0.0 & 0.0 & 977 & 99.5 & 5   & 0.5    & 0	 & 0.0 \\
\multirow{-2}{*}{Mistral-22B}    & one  & 632 & 99.8 & 1   & 0.2	& 0.0 & 0.0 & 962 & 98.0 & 20  & 2.0    & 0	 & 0.0 \\
& zero & 582 & 91.9 & 47  & 7.4	& 4.0 & 0.6 & 881 & 89.7 & 101 & 10.3   & 0	 & 0.0 \\
\multirow{-2}{*}{Olmo2-32B}      & one  & 627 & 99.1 & 6   & 0.9	& 0.0 & 0.0 & 893 & 90.9 & 89  & 9.1    & 0	 & 0.0 \\
& zero & 616 & 97.3 & 17  & 2.7	& 0.0 & 0.0 & 904 & 92.1 & 78  & 7.9    & 0	 & 0.0 \\
\multirow{-2}{*}{Mixtral-8x22B}  & one  & 604 & 95.4 & 29  & 4.6	& 0.0 & 0.0 & 919 & 93.6 & 60  & 6.1    & 3	 & 0.3 \\ \midrule
& zero & 617 & 97.5 & 14  & 2.2	& 2.0 & 0.3 & 932 & 94.9 & 46  & 4.7    & 4	 & 0.4 \\
\multirow{-2}{*}{Llama-70B}      & one  & 606 & 95.7 & 27  & 4.3	& 0.0 & 0.0 & 936 & 95.3 & 44  & 4.5    & 2	 & 0.2 \\
& zero & 607 & 95.9 & 24  & 3.8	& 2.0 & 0.3 & 915 & 93.2 & 66  & 6.7    & 1	 & 0.1 \\
\multirow{-2}{*}{Qwen2.5-72B}    & one  & 612 & 96.7 & 21  & 3.3	& 0.0 & 0.0 & 920 & 93.7 & 61  & 6.2    & 1	 & 0.1 \\
& zero & 562 & 88.8 & 70  & 11.1	& 1.0 & 0.2 & 688 & 70.1 & 288 & 29.3   & 6	 & 0.6 \\
\multirow{-2}{*}{Com.-R-104B}    & one  & 615 & 97.2 & 18  & 2.8	& 0.0 & 0.0 & 862 & 87.8 & 114 & 11.6   & 6	 & 0.6 \\ \bottomrule
\end{tabular}}
\caption{Variability across three runs per LLM in the check-worthiness annotation task. It reports the proportion of cases where all three runs produced the same label (\text{all equal}), two out of three runs agreed (2 equal), and all runs produced different labels (\text{unequal}).}
\label{tab:variability}
\end{table}

\begin{table*}[t]
\resizebox{!}{1.133cm}{\
\begin{tabular}{lllllllllllllllllll}
\toprule
\textbf{} & \multicolumn{2}{l}{ALL}        & \multicolumn{2}{l}{DISABLED}  & \multicolumn{2}{l}{JEWS}      & \multicolumn{2}{l}{LGBT+}     & \multicolumn{2}{l}{MIGRANTS}  & \multicolumn{2}{l}{MUSLIMS}   & \multicolumn{2}{l}{POC}       & \multicolumn{2}{l}{WOMEN}     & \multicolumn{2}{l}{other}     \\ \midrule
                                  & \#             & \%              & \#            & \%              & \#            & \%              & \#            & \%              & \#            & \%              & \#            & \%              & \#            & \%              & \#            & \%              & \#            & \%              \\ \midrule
CFS	& 2128          & 42.53          & 62             & 28.18          & \textbf{387}   & \textbf{65.15} & 166          & 26.90          & \textbf{517}	& \textbf{54.02}    & 620          & 46.44          & 127	        & 36.08         	& 150          & 22.66          & 99	        & 37.22         \\
NFS	& \textbf{2766} & \textbf{55.29} & \textbf{148}   & \textbf{67.27} & 193            & 32.49          & \textbf{441} & \textbf{71.47} & 427	        & 44.62	            & \textbf{695} & \textbf{52.06} & \textbf{207}	& \textbf{58.81}	& \textbf{498} & \textbf{75.23} & \textbf{157}	& \textbf{59.02} \\
UFS	& 109           & 2.179          & 10             & 4.55           & 14             & 2.36           & 10           & 1.62           & 13	        & 1.36              & 20           & 1.50           & 18            & 5.11	            & 14           & 2.11           & 10	        & 3.76            \\ \bottomrule
\end{tabular}} \caption{Olmo2-32B zero-shot check-worthiness analysis on the MT-CONAN dataset, where all messages are HS, reported both overall and disaggregated by target group.} \label{tab:mt-conan-cw}
\end{table*}

\section{Prompts for the Hate Speech Detection Task}
\label{sec:add-details-hs-misinfo}


For the hate speech detection task described in Section \ref{sec:results-hs-det-and-openaitool}, rather than classifying individual claims within messages, we performed classification at the message level. Messages—annotated as either HS or Non-HS) from WSF-ARG+—were classified using a system-user prompt with structured outputs:

\noindent \textbf{System Prompt:} \texttt{You are an expert model for detecting hate speech in text messages based on the following definition of hate speech: Hate speech is considered any kind of content that conveys malevolent intentions toward a group or an individual, and motivated by inherent characteristics that are attributed to that group and shared among its members such as race, color, ethnicity, gender, sexual orientation, nationality, religion, disability, social status, health conditions, or other characteristics.}

\noindent \textbf{User Prompt:} \texttt{Given this message. Classify if it is either hateful or not. \\ Input: \{input\_text\}
}

\noindent \textbf{Output Labels:} \texttt{["hateful", "non-hateful"]}

We conducted experiments on two groups: one with check-worthy labels and one without. In the configuration without check-worthy labels, the \texttt{input\_text} consisted of the full HS or Non-HS message to be classified. In the configuration with check-worthy labels, the \texttt{input\_text} included the full HS or Non-HS message, but each of the claims composing the message where wrapped with their corresponding check-worthiness annotations. For comparison among the two groups, instead of using the original entire messages in WSF-ARG+, we used the concatenation of the claims that compose them. For example, the following hate speech message consisting of two claims (claim 1 represented with \ctext[RGB]{255,187,158}{one color}, and claim 2 represented with \ctext[RGB]{181,228,224}{another color}):

\noindent \textbf{W/o Check-worthiness}: \textit{\ctext[RGB]{255,187,158}{Before the age of 10, I had never seen a black person in real life.} \ctext[RGB]{181,228,224}{I was horrified by the amount of non-white people after moving to Holland.}}

\noindent \textbf{W/ Check-worthiness}: \textit{\texttt{\textbf{[Unimportant Factual]}} \ctext[RGB]{255,187,158}{Before the age of 10, I had never seen a black person in real life.} \texttt{\textbf{[/Unimportant Factual]}} \texttt{\textbf{[Check-worthy Factual]}} \ctext[RGB]{181,228,224}{I was horrified by the amount of non-white people after moving to Holland.} \texttt{\textbf{[/Check-worthy Factual]}}}



\section{Sociodemographics of the Annotators}
\label{sec:annotators-demographics}

In total, five annotators participated across the different annotation tasks. All annotators are researchers with backgrounds in computer science, computational linguistics, or natural language processing, and have prior familiarity with hate speech research. The group consists of one Postdoctoral researcher, two PhD students, and two senior researchers. Annotators are from Latin American and European countries and were based in Europe at the time of the study. The age-range of the annotators are 26-45. All annotators have bilingual proficiency in English. Annotators received written instructions and were trained on the annotation guidelines prior to the annotation process. Participation in the annotation tasks was voluntary.

\paragraph{Claim Identification Annotators in Non-Hate Speech Messages}
Two annotators were involved in this task: one Postdoctoral researcher and one PhD student in Computer Science. One annotator identified as male and the other as non-binary. One annotator is from a Latin American country and the other from a European country. We use the same instructions described by \citet{bonaldi-etal-2024-safer} for the identification of claims in HS messages.

\paragraph{Check-worthiness Annotators for the LLM-in-the-loop Framework}
Two annotators participated in this task: one Postdoctoral researcher and one PhD student in Computer Science. One annotator identified as male and the other as female. Both annotators are from Latin American countries. We use the definitions and instructions from \citet{claimbuster}. The annotation guidelines are available in the repository of the WSF-ARG+ dataset.

\paragraph{Check-worthiness Annotators for the Full Human Approach}
Four annotators participated in this task. The group included one Postdoctoral researcher, one PhD student, and two senior researchers from Latin American and European countries. One identified as non-binary and the other three annotators as male. Instructions and guidelines are the same used for the LLM-in-the-loop framework.

\begin{table*}[t]
\resizebox{!}{3.62cm}{\
\begin{tabular}{lccc|ccc}
\toprule
\multirow{2}{*}{Model} & \multicolumn{3}{c|}{Zero-Shot}                                                                                                                                                                                                    & \multicolumn{3}{c}{One-Shot}                                                                                                                                                                                                     \\
                       & \multicolumn{1}{c}{Time (Sec)} & \multicolumn{1}{c}{Power (W)} & \multicolumn{1}{c|}{Energy (kWh)} & \multicolumn{1}{c}{Time (Sec)} & \multicolumn{1}{c}{Power (W)} & \multicolumn{1}{c}{Energy (kWh)} \\
\midrule
Mistral-7B             & 93.00   $\pm$ 17.52   & 434.59  $\pm$ 220.38  & 0.011 $\pm$ 0.005 & 76.33  $\pm$ 8.62    & 280.11  $\pm$ 157.08 & 0.006 $\pm$ 0.003  \\
Llama-8B               & 76.33   $\pm$ 3.79    & 400.28  $\pm$ 93.40   & 0.008 $\pm$ 0.002 & 71.33  $\pm$ 19.43   & 352.43  $\pm$ 229.16 & 0.006 $\pm$ 0.003  \\
Olmo-7B                & 79.67   $\pm$ 8.33    & 342.56  $\pm$ 171.52  & 0.008 $\pm$ 0.004 & 75.33  $\pm$ 5.03    & 267.92  $\pm$ 139.32 & 0.005 $\pm$ 0.002  \\
Qwen2.5-7B             & 77.33   $\pm$ 12.50   & 359.08  $\pm$ 182.83  & 0.008 $\pm$ 0.004 & 66.33  $\pm$ 13.43   & 319.80  $\pm$ 77.42  & 0.006 $\pm$ 0.002  \\
Command-R-7B           & 80.67   $\pm$ 4.93    & 275.22  $\pm$ 131.28  & 0.006 $\pm$ 0.003 & 64.67  $\pm$ 10.12   & 253.61  $\pm$ 47.11  & 0.005 $\pm$ 0.001  \\
\midrule
Avg. Small             & 81.40   $\pm$ 9.41    & 362.35  $\pm$ 159.88  & 0.008 $\pm$ 0.004 & 70.80  $\pm$ 11.32   & 294.77  $\pm$ 130.02 & 0.006 $\pm$ 0.002  \\
\midrule
Mixtral-8x7B           & 341.33  $\pm$ 27.15   & 921.61  $\pm$ 53.55   & 0.088 $\pm$ 0.011 & 315.67 $\pm$ 35.57   & 936.21  $\pm$ 17.37  & 0.082 $\pm$ 0.010  \\
Mistral-22B            & 148.33  $\pm$ 39.83   & 258.88  $\pm$ 68.11   & 0.011 $\pm$ 0.006 & 124.33 $\pm$ 4.16    & 374.94  $\pm$ 98.57  & 0.013 $\pm$ 0.004  \\
Olmo2-32B              & 155.67  $\pm$ 6.11    & 359.09  $\pm$ 77.04   & 0.016 $\pm$ 0.004 & 150.00 $\pm$ 5.29    & 331.05  $\pm$ 53.53  & 0.014 $\pm$ 0.002  \\
Mixtral-8x22B          & 692.00  $\pm$ 89.71   & 969.52  $\pm$ 29.27   & 0.187 $\pm$ 0.028 & 555.67 $\pm$ 374.51  & 1014.66 $\pm$ 94.46  & 0.150 $\pm$ 0.096  \\
\midrule
Avg. Medium            & 334.33  $\pm$ 40.70   & 627.27  $\pm$ 56.99   & 0.075 $\pm$ 0.012 & 286.42 $\pm$ 104.89  & 664.21  $\pm$ 65.98  & 0.065 $\pm$ 0.028  \\
\midrule
Llama-70B              & 377.33  $\pm$ 32.65   & 961.52  $\pm$ 13.32   & 0.101 $\pm$ 0.008 & 302.33 $\pm$ 166.63  & 1025.76 $\pm$ 38.25  & 0.085 $\pm$ 0.046  \\
Qwen2.5-72B            & 397.67  $\pm$ 36.12   & 969.18  $\pm$ 55.44   & 0.107 $\pm$ 0.015 & 206.67 $\pm$ 171.13  & 1049.80 $\pm$ 30.61  & 0.060 $\pm$ 0.048  \\
Command-R-104B         & 480.33  $\pm$ 35.81   & 985.00  $\pm$ 46.03   & 0.132 $\pm$ 0.015 & 256.33 $\pm$ 217.60  & 1055.96 $\pm$ 129.00 & 0.072 $\pm$ 0.055  \\
\midrule
Avg. Large             & 418.44  $\pm$ 34.86   & 971.90  $\pm$ 38.26   & 0.113 $\pm$ 0.013 & 255.11 $\pm$ 185.12  & 1043.84 $\pm$ 65.95  & 0.072 $\pm$ 0.050 \\ 
\bottomrule
\end{tabular}}
\caption{Average energy consumption and standard deviation (±) of LLMs in the zero-shot and one-shot settings for the check-worthiness detection task, measured across three independent runs. We report execution time in seconds (s), average power in watts (W), and total energy consumption in kilowatt-hours (kWh).}
\label{tab:energy-llms-cw}
\end{table*}

\section{Check-worthiness Detection in Other HS Datasets Using Olmo2-32B}
\label{sec:cw-det-other-datasets}

To analyze the frequency of check-worthy claims beyond the WSF-ARG+ dataset, we applied it to MT-CONAN, which is designed to generate fact-based counter-narratives for HS messages. In this context, since the primary goal is to produce factual responses, HS messages should at least contain claims that are worth debating, i.e., check-worthy. We ran the best model configuration that achieved the highest agreement with humans for the LLM-in-the-loop evaluation, Olmo2-32B. Table \ref{tab:mt-conan-cw} presents the number of CFS, NFS, and UFS annotations for HS messages in MT-CONAN. The results show that, even in a dataset focused on fact-based counter-speech, only 42.53\% of the messages are actually CFS. This pattern is consistent across nearly all target groups.

\section{LLM Energy Consumption}
\label{sec:energy-llms}

\begin{table*}[t]
\small
\centering
\resizebox{!}{3.4cm}{\
\begin{tabular}{lccc|ccc}
\toprule
\multirow{2}{*}{Model} & \multicolumn{3}{c|}{w/o check-worthiness}                                                                                                                                                                                         & \multicolumn{3}{c}{w/ check-worthiness}                                                                                                                                                                                          \\
                       & \multicolumn{1}{c}{Time (Sec)} & \multicolumn{1}{c}{Power (W)} & \multicolumn{1}{c|}{Energy (kWh)} & \multicolumn{1}{c}{Time (Sec)} & \multicolumn{1}{c}{Power (W)} & \multicolumn{1}{c}{Energy (kWh)}  \\
\midrule
Mistral-7B             & 80.67   $\pm$ 9.50    & 406.17   $\pm$ 234.58  & 0.009  $\pm$ 0.005  & 76.67   $\pm$ 3.79     & 442.47  $\pm$ 58.04    & 0.009  $\pm$ 0.002  \\
Llama-8B               & 79.00   $\pm$ 1.73    & 672.88   $\pm$ 459.12  & 0.015  $\pm$ 0.010  & 70.67   $\pm$ 18.01    & 460.55  $\pm$ 302.52   & 0.008  $\pm$ 0.003  \\
Olmo-7B                & 74.33   $\pm$ 2.52    & 435.66   $\pm$ 113.14  & 0.009  $\pm$ 0.002  & 67.33   $\pm$ 6.66     & 668.82  $\pm$ 247.86   & 0.012  $\pm$ 0.004  \\
Qwen2.5-7B             & 74.00   $\pm$ 9.54    & 795.53   $\pm$ 141.25  & 0.017  $\pm$ 0.005  & 64.33   $\pm$ 13.61    & 570.32  $\pm$ 303.05   & 0.010  $\pm$ 0.004  \\
Command-R-7B           & 79.00   $\pm$ 14.00   & 444.17   $\pm$ 271.20  & 0.009  $\pm$ 0.006  & 69.67   $\pm$ 7.51     & 262.15  $\pm$ 33.50    & 0.005  $\pm$ 0.000  \\
\midrule
Avg. Small             & 77.40   $\pm$ 7.46    & 550.88   $\pm$ 243.86  & 0.012  $\pm$ 0.006  & 69.73   $\pm$ 9.91     & 480.86  $\pm$ 188.99   & 0.009  $\pm$ 0.003  \\
\midrule
Mixtral-8x7B           & 425.00  $\pm$ 29.61   & 927.90   $\pm$ 32.89   & 0.110  $\pm$ 0.009  & 438.33  $\pm$ 13.65    & 935.44  $\pm$ 41.59    & 0.114  $\pm$ 0.009  \\
Mistral-22B            & 132.00  $\pm$ 9.54    & 407.46   $\pm$ 216.87  & 0.015  $\pm$ 0.007  & 129.67  $\pm$ 3.79     & 428.94  $\pm$ 72.41    & 0.015  $\pm$ 0.002  \\
Olmo2-32B              & 152.67  $\pm$ 10.97   & 453.85   $\pm$ 51.28   & 0.019  $\pm$ 0.003  & 154.67  $\pm$ 7.37     & 386.20  $\pm$ 178.14   & 0.016  $\pm$ 0.007  \\
Mixtral-8x22B          & 911.00  $\pm$ 27.84   & 953.39   $\pm$ 17.55   & 0.241  $\pm$ 0.008  & 644.33  $\pm$ 449.61   & 1091.45 $\pm$ 193.88   & 0.180  $\pm$ 0.118  \\
\midrule
Avg. Medium            & 405.17  $\pm$ 19.49   & 685.65   $\pm$ 79.65   & 0.096  $\pm$ 0.007  & 341.75  $\pm$ 118.61   & 710.51  $\pm$ 121.50   & 0.081  $\pm$ 0.034  \\
\midrule
Llama-70B              & 468.00  $\pm$ 9.17    & 940.11   $\pm$ 24.63   & 0.122  $\pm$ 0.005  & 447.00  $\pm$ 10.54    & 981.88  $\pm$ 54.74    & 0.122  $\pm$ 0.010  \\
Qwen2.5-72B            & 368.67  $\pm$ 189.89  & 978.25   $\pm$ 42.36   & 0.100  $\pm$ 0.052  & 465.67  $\pm$ 15.50    & 973.42  $\pm$ 27.84    & 0.126  $\pm$ 0.006  \\
Command-R-104B         & 430.00  $\pm$ 278.54  & 1001.50  $\pm$ 20.85   & 0.120  $\pm$ 0.078  & 413.33  $\pm$ 267.05   & 1062.16 $\pm$ 148.38   & 0.115  $\pm$ 0.068  \\
\midrule
Avg. Large             & 422.22  $\pm$ 159.20  & 973.29   $\pm$ 29.28   & 0.114  $\pm$ 0.045  & 442.00  $\pm$ 97.70    & 1005.82 $\pm$ 76.98    & 0.121  $\pm$ 0.028 \\
\bottomrule
\end{tabular}}
\caption{Average energy consumption and standard deviation (±) of LLMs without (w/o) check-worthiness and with (w/) check-worthiness (as shown in the prompts of Section~\ref{sec:add-details-hs-misinfo}) for the hate speech detection task, measured across three independent runs. We report execution time in seconds (s), average power in watts (W), and total energy consumption in kilowatt-hours (kWh).}
\label{tab:energy-llms-hs}
\end{table*}

We calculated the energy consumption of our experiments for both check-worthiness detection and hate speech detection. Mistral-22B, Olmo2-32B, and all small models were run on a single NVIDIA H100 GPU. Mixtral-8x7B, Mixtral-8x22B, and all large models were run on four NVIDIA H100 GPUs. Computations were done in Snellius, the Dutch National supercomputer hosted at SURF~\footnote{\url{https://www.surf.nl/en/services/compute/snellius-the-national-supercomputer}}. Estimations were done using the Energy Aware Runtime (EAR) package~\footnote{\url{https://gitlab.bsc.es/ear_team/ear}}, an energy management framework for supercomputers.


For check-worthiness detection, Table~\ref{tab:energy-llms-cw} reports the average energy consumption of the LLMs under zero-shot and one-shot settings across the three independent runs performed for the LLM-in-the-loop evaluation described in Section~\ref{sec:experimental-settings}. We measured execution time in seconds (Sec.), power in watts (W), and total energy consumption in kilowatt-hours (kWh). To annotate all 1615 claims in WSF-ARG+, small-size LLMs ranged from 64.67 to 93.00 seconds and from 253.61 to 434.59 W, medium-size LLMs ranged from 124.33 to 692.00 seconds and from 331.05 to 1014.66 W, and large-size LLMs ranged from 206.67 to 480.00 seconds and from 961.52 to 1055.96 W. Mixtral-8x7B and Mixtral-8x22B exhibited execution time and power comparable to large-size LLMs, likely due to their Mixture of Experts architecture and the higher GPU memory required to load their model weights. However, the number of claims annotated in this experiment is relatively small, and the efficiency advantages of Mixture of Experts of these models at prediction time may become more apparent on larger datasets. Overall, small-size models consumed an average of 0.008 kWh in the zero-shot setting and 0.006 kWh in the one-shot setting, medium-size models consumed 0.075 kWh and 0.065 kWh on average, and large-size models consumed 0.113 kWh and 0.072 kWh on average for zero-shot and one-shot, respectively. The configuration with the highest energy consumption was Mixtral-8x22B in the zero-shot setting, reaching 0.187 kWh. Execution time, power, and energy consumption were lower in the one-shot setting than in the zero-shot setting for all models except Mistral-22B, although the differences between the two prompting configurations were generally small. Initially, we hypothesized that the higher number of input tokens in the one-shot configuration would result in higher energy consumption; however, this was not observed in our experiments. Still, while there was a lower energy consumption for one-shot on average, it presents a higher variability than the runs in zero-shot. Across all three runs, LLMs, and prompts, the total execution time was 15794 seconds (4.39 hours), with a cumulative power draw of 43496.36 W and a total energy consumption of 3.55 kWh.

For hate speech detection, Table~\ref{tab:energy-llms-hs} shows the energy consumption of LLMs instructed for hate speech detection with and without platinum check-worthiness annotations. A similar pattern to the one observed for check-worthiness can be seen. On average, to annotate all 521 messages in WSF-ARG+, small models consumed 0.012 kWh without check-worthiness and 0.009 kWh with check-worthiness, while medium-sized models consumed 0.096 kWh and 0.081 kWh, respectively. Large models showed the highest consumption, at 0.114 kWh without check-worthiness and 0.121 kWh with check-worthiness. Higher energy consumption was required for small and medium-sized models under the without check-worthiness runs, whereas higher consumption was observed for large models under the with check-worthiness configuration, which was also the configuration that benefited from including these annotations in the prompt to predict hatefulness, as shown in Table~\ref{tab:hs-clf-scores-with-std}. For medium-sized models, Mixtral-8x7B and Mixtral-8x22B consumed even more energy than the large models, reaching up to 0.241 kWh on average, almost twice the amount consumed by large-sized models. Overall, across all three runs and all evaluated LLMs, the experiments without check-worthiness labels required a total execution time of 9823 seconds (2.73 hours), with a cumulative power draw of 25250.61 W, corresponding to an energy consumption of 2.36 kWh. The experiments with check-worthiness using platinum labels required 9125 seconds (2.53 hours), with a cumulative power draw of 24791.41 W, corresponding to an energy consumption of 2.20 kWh. Since we also ran the experiments with gold labels (as shown in Table~\ref{tab:hs-clf-scores-with-std}), we estimate their energy consumption to be approximately equivalent to that of the platinum-label experiments. Thus, the total energy consumption across the three configurations—hate speech detection without check-worthiness labels, with gold check-worthiness labels, and with platinum check-worthiness labels—is estimated as $2.36~\text{kWh} + 2 \times 2.20~\text{kWh} = 6.76~\text{kWh}$.

The best configuration for check-worthiness detection was the medium-sized Olmo2-32B model under the zero-shot setting (see details in Table~\ref{tab:agreement}). On average, it consumed 0.016 kWh to annotate the claims in WSF-ARG+. In contrast, the most performant LLM for hate speech detection was Olmo2-7B with gold annotations, achieving an average F1 score of 0.776 while consuming 0.008 kWh to annotate all messages in WSF-ARG+. These results suggest that, at least for these two tasks and on this dataset, opting for smaller, more energy-efficient models may be preferable to using larger models. 

Summing the two tasks, the estimated total energy consumption required to run all our LLM experiments in the paper was 10.31 kWh. Using The Netherlands conversion factor for carbon dioxide-equivalents emitted per kilowatt-hour of electricity in 2025, 253.56 gCO$_2$eq per kWh~\cite{ember2026_lifecycle_carbon_intensity}, we estimate the paper's carbon footprint as 10.31 kWh $\times$ 253.56 $\frac{\text{gCO}_2\text{eq}}{\text{kWh}}$ $\approx$ 2.61 kgCO$_2$eq. 

\end{document}